\documentclass[12pt]{article}

\usepackage{microtype,enumerate}
\usepackage{graphicx}
\usepackage{subfigure}
\usepackage{booktabs} 
\usepackage{makecell,diagbox}
\usepackage{siunitx}
\usepackage{amssymb}

\usepackage{algorithm,algorithmic}
\usepackage{bm}
\usepackage[numbers]{natbib}
\usepackage{graphicx}
\usepackage{amssymb,amsmath}

\oddsidemargin
-0.2in \topmargin -.60in
\usepackage{mathtools}
\usepackage{amsthm}
\usepackage{diagbox}
\usepackage{multirow}
\usepackage[capitalize,noabbrev]{cleveref}
\usepackage{threeparttable}
\usepackage{threeparttablex}

\begin{document}

\title{Revisiting PINNs: Generative Adversarial Physics-informed Neural Networks and Point-weighting Method}

\author{
	Wensheng~Li$^{1}$,  Chao~Zhang$^{1}$\footnote{Corresponding author.}, Chuncheng Wang$^{2}$, Hanting Guan$^{1}$, Dacheng Tao$^{3}$\\
	$^1$School of Mathematical Sciences, Dalian University of Technology, China \\
	\texttt{Lws@mail.dlut.edu.cn; chao.zhang@dlut.edu.cn; miracie@mail.dlut.edu.cn} \\
	$^{2}$School of Mathematics, Harbin Institute of Technology, China\\
	\texttt{wangchuncheng@hit.edu.cn} \\
	$^{3}$JD Explore Academy, China \\
	\texttt{dacheng.tao@gmail.com} \\
}

\maketitle
\begin{abstract}

Physics-informed neural networks (PINNs) provide a deep learning framework for numerically solving partial differential equations (PDEs), and have been widely used in a variety of PDE problems. However, there still remain some challenges in the application of PINNs: 1) the mechanism of PINNs is unsuitable (at least cannot be directly applied) to exploiting a small size of (usually very few) extra informative samples to refine the networks; and 2) the efficiency of training PINNs often becomes low for some complicated PDEs. In this paper, we propose the generative adversarial physics-informed neural network (GA-PINN), which integrates the generative adversarial (GA) mechanism with the structure of PINNs, to improve the performance of PINNs by exploiting only a small size of exact solutions to the PDEs. Inspired from the weighting strategy of the Adaboost method, we then introduce a point-weighting (PW) method to improve the training efficiency of PINNs, where the weight of each sample point is adaptively updated at each training iteration. The numerical experiments show that GA-PINNs outperform PINNs in many well-known PDEs and the PW method also improves the efficiency of training PINNs and GA-PINNs.

\end{abstract}

\section{Introduction} \label{sec:introduction}

Partial differential equations (PDEs) are the core mathematical tools for studying the evolvement of physical systems, and play an essential role in various engineering applications, {\it e.g.,} Navier-Stokes equations for hydrodynamics \citep{ershkov2021towards} and Laplace's equation for celestial problems \citep{orti2022alternative}. Consider a general form of partial differential equations (PDEs): 

\begin{equation}\label{eq:pde}
	\left\{
	\begin{array}{l l}
		\mathcal{L}[u]({\bf x})=F({\bf x}), &    {\bf x} \in \Omega;  \\
		\mathcal{B}_i[u]({\bf x})= G_i ({\bf x}), &   {\bf x} \in \partial_i \Omega,\;\; 1\leq i\leq I,
	\end{array}
	\right. 
\end{equation}

where $u({\bf x}): \Omega \rightarrow \mathbb{R}^d$ is the solution to the PDE problem; $\mathcal{L}$ is the nonlinear partial differential operator; $\mathcal{B}_i$ is the boundary condition operator; $\Omega\subset\mathbb{R}^D$ is the domain of PDE problem; and $\partial_i \Omega$ are the boundaries of the domain $\Omega$ ($1\leq i\leq I$). Since the operators $\mathcal{L}$ and $\mathcal{B}_i$ usually have complicated forms, it is difficult to obtain the analytical solutions to PDEs. Instead, the numerical methods become the main manners for solving PDEs, {\it e.g.,} finite element method (FEM) \citep{zienkiewicz1977finite}, finite difference method (FDM) \citep{smith1985numerical} and finite volume method (FVM) \citep{leveque2002finite}. However, especially for complex PDEs, these methods often need high computational cost and extra expert knowledge, {\it e.g.,} the mesh generation in FEM, the setting of complicated boundaries in FDM, and the treatment of irregular geometries in FVM.


\subsection{Physics-informed Neural Networks}\label{sec:pinn}

Physics-informed neural networks (PINNs) provide a deep learning framework for numerically solving PDEs based on the view that if a neural network $\widehat{u}({\bf x})$ holds the equation and the boundary conditions of the PDE problem \eqref{eq:pde}, it will be an accurate approximation to the exact solution $u({\bf x})$ \citep{dissanayake1994neural,raissi2019physics}. To guarantee the differentiability of $\widehat{u}({\bf x})$, all activation functions in the network are set to be $\tanh(\cdot)$. The equation loss ${\rm L}_f$ and the $i$-th boundary-condition loss ${\rm L}_{b_i}$ are defined as follows: 

\begin{equation*}
	\begin{array}{l}
		{\rm L}_f:=  \frac{1}{N} \sum_{n=1}^{N} \big\|\mathcal{L}[\widehat{u}]\big({\bf x}^{(n)}\big)-f\big({\bf x}^{(n)}\big)\big\|^{2}, \\
		{\rm L}_{b_i}:=  \frac{1}{M_{i}} \sum_{m=1}^{M_{i}}\big\|\mathcal{B}_i[\widehat{u}]\big({\bf x}^{(m)}_{i}\big)-g_i\big({\bf x}^{(m)}_{i}\big)\big\|^{2}.
	\end{array} 
\end{equation*}

Then, the weights of $\widehat{u}({\bf x})$ can be empirically obtained by minimizing the physics-informed (PI) loss:
\begin{equation}\label{eq:MSE}
	\begin{array}{l}
		{\rm L}_{\rm PINN}  :=  {\rm L}_f + \lambda_1 {\rm L}_{b} \;\;  \mbox{with} \;\;{\rm L}_{b} := \sum_{i=1}^I {\rm L}_{b_i},
	\end{array} 
\end{equation}

where ${\rm L}_{b}$ is called the boundary-condition loss; $\lambda_1>0$ is a constant to adjust the proportion between the ${\rm L}_f$ and ${\rm L}_{b}$; $\{{\bf x}^{(n)}\}_{n=1}^{N}$ are the points taken from the domain $\Omega$; and $\{{\bf x}_{i}^{(m)}\}_{m=1}^{M_{i}}$ are the points taken from the boundaries $\partial_i \Omega$ of domain ($1\leq i\leq I$). Take Poisson equation as an example:
\begin{equation}\label{eq:Possion}
	\left\{
	\begin{array}{l l}
		\Delta u =-\sin (\pi x) \sin (\pi y), &(x, y) \in \Omega; \\
		u (x,0)=u (x,1)=0, &(x, y) \in \partial_1 \Omega ; \\
		u (0,y)=u (1,y)=0, &(x, y) \in \partial_2 \Omega ,
	\end{array}
	\right. 
\end{equation}

where $\Omega = (0,1) \times(0,1)$, $\partial_1 \Omega = (0,1)\times\{0,1\}$ and $\partial_2 \Omega = \{0,1\}\times (0,1)$. Let $\widehat{u}({\bf x})$ be the network providing numerical solutions to the PDE and its weights are obtained by minimizing the PI loss ${\rm L}_{\rm PINN}:={\rm L}_{f}+\lambda_1\left({\rm L}_{b_1}+{\rm L}_{b_2}\right)$
%
with
\begin{align*}
	&{\rm L}_{f}=\frac{1}{N} \sum_{n=1}^{N}\Big[\Delta \widehat{u}(x^{(n)}, y^{(n)})+\sin (\pi x^{(n)}) \sin (\pi y^{(n)})\Big]^{2}; \\
	&{\rm L}_{b_1}=\frac{1}{M_{1}}\sum_{m=1}^{M_{1}}\left[\widehat{u}\big(x_{1}^{(m)}, 0\big)\right]^{2}+\left[\widehat{u}\big(x_{1}^{(m)}, 1\big)\right]^{2}; \\
	&{\rm L}_{b_2}=\frac{1}{M_{2}}\sum_{m=1}^{M_{2}}\left[\widehat{u}\big(0,y_{2}^{(m)}\big)\right]^{2}+\left[\widehat{u}\big(1,y_{2}^{(m)}\big)\right]^{2}, 
\end{align*}
where $(x^{(n)}, y^{(n)})$ are taken from the domain $\Omega$; $(x_{1}^{(m)}, 0)$ and $(x_{1}^{(m)}, 1)$ (resp. $(0,y_{2}^{(m)})$ and $(1,y_{2}^{(m)})$) are taken from the boundary $\partial_1 \Omega$ (resp. $\partial_2 \Omega$).

Different from the traditional PDE numerical solution methods, PINNs only need to build a network with an appropriate structure and then to determine the network weights by minimizing ${\rm L}_{\rm PINN}$ rather than to consider the mathematical properties of the PDEs. Because of the high applicability, PINNs have been widely used to solve a variety of PDEs, {\it e.g.,} the Euler equations for high-speed aerodynamic flows simulation \citep{mao2020physics} and the Fokker-Planck equation for inverse stochastic problems from discrete particle observations \citep{chen2021solving}. Moreover, \citet{wang2022and} adopted the neural tangent kernel method to study the reason why PINNs sometimes cannot be trained. \citet{shin2020convergence} studied the convergence of generalization error of PINNs for linear second-order elliptic and parabolic PDEs. \citet{mishra2021estimates} estimated the generalization error of PINNs based on the training error and the size of training samples. In addition, there are also some variants of PINNs, such as MPINN for multi-fidelity surrogate modeling \citep{meng2020composite} and B-PINNs for quantifying the aleatoric uncertainty arising from the noisy data \citep{yang2021b}.


However, there are some caveats in the applications of PINNs:
\begin{enumerate}[(1)]

	\item To maintain a strong nonlinear mapping capability, the network $\widehat{u}({\bf x})$ usually has multiple hidden layers and a large amount of nodes in each hidden layer. Such a complicated structure significantly increases the training difficulty. 
	
	%

	\item Different from the general supervised learning setting, the sample points used to train PINNs are only taken from the domain $\Omega$ and its boundaries $\partial_i \Omega$. There is no labeled sample of the form $({\bf x}^{(n)}, u({\bf x}^{(n)}))$ (${\bf x}^{(n)}\in\Omega$) to correct the network outputs.\footnote{As addressed by \citet{raissi2019physics}, in some cases, the boundary-condition loss ${\rm L}_b$ in PINNs can be replaced with the mean squared loss computed on a small size of labeled boundary samples $\{({\bf x}_i^{(m)}, u({\bf x}_i^{(m)}))\}_{m=1}^{M_i}$ (${\bf x}_i^{(m)}\in\partial_i\Omega$).} Therefore, the process of training PINNs often becomes unstable and hard to control. 

	\item Let $\{({\bf x}_{T}^{(j)},u_T^{(j)})\}_{j=1}^J$ be a set of labeled samples, which are the exact solutions $u_T^{(j)}$ to the PDE at the points ${\bf x}_T^{(j)}$. This situation is common in practice. For example, some sensors are equipped on the surface of a physical objective to capture its real-time physical responses, or some numerical methods can provide high-accuracy approximations of the exact solutions at some points in the domain. Following the mechanism of PINNs, a natural way of exploiting these labeled samples is to minimize the objective function  
	\begin{equation}\label{eq:MSE2}
		\overline{\rm L}_{\rm PINN}:={\rm L}_{\rm PINN} + \lambda_2  {\rm L}_T, 
	\end{equation}
	where $ {\rm L}_T = \frac{1}{J} \sum_{j=1}^J \| \widehat{u}({\bf x}_T^{(j)})  - u_T^{(j)} \|^2$. Unfortunately, when the labeled sample size $J$ is small, the minimization of ${\rm L}_T$ is likely to bring a negative effect to the training performance provided by the minimization of ${\rm L}_{\rm PINN}$.


\end{enumerate}

\subsection{Generative Adversarial Networks}

Generative adversarial networks (GANs), proposed by \citet{goodfellow2014generative}, are referred to a class of neural networks that are composed of two sub-networks: the generator and the discriminator, and are trained in the manner of minimax game between them. Taking advantage of the specific network structure with the minimax game training, many empirical evidences have shown that GANs can approximate the data distribution accurately with a lower demand on the sample size than the traditional machine-learning models. This characteristic provides an applicable way of exploiting a limited size of labeled samples to refine the networks. For example, some variants of GANs have been successfully used to deal with multi-fidelity surrogate modeling \citep{liu2021gan} and the TBM tunnel geological prediction \citep{zhang2022generative}, where only a small size of labeled samples are available. Such a characteristic motivates this paper as well.

\subsection{Overview of Main Results}

In this paper, we mainly concern with two issues on PINNs: one is how to improve the training efficiency and the other is how to exploit a small size of (usually very few) labeled samples to refine the networks. 

Inspired from the weighting strategy of the Adaboost method \citep{freund1997decision}, we introduce a point-weighting (PW) method to improve the training performance of PINNs, where the weight of every sample point is adaptively updated at each training iteration. This method splits the process of training PINNs into two stages. In the first stage, since the network is relatively fragile, the weights of points that provide small (resp. big) training errors are increased (resp. decreased) to stabilize the network as quickly as possible. In the second stage, the weights of the points that have big (resp. small) training errors are increased (resp. decreased) to make the network  fit all sample points accurately. The numerical experiments show that the proposed PW method not only speeds up the training process but also improves the network performance.

We then consider the extension of PINNs in the situation that there are a small size of (usually very few) labeled samples, {\it i.e.,} the exact solutions or their high-accuracy approximations at some discrete points in the domain. As stated above, it is difficult for the mechanism of PINNs to handle this situation. Instead, by integrating the generative adversarial (GA) mechanism with the structure of PINNs, we propose the generative adversarial physics-informed neural networks (GA-PINNs) to numerically solve PDEs in this situation.


The GA-PINN is composed of two sub-networks: a generator and a discriminator. The generator is a PINN, {\it i.e.,} a network that, given an input ${\bf x}\in\Omega$, produces the approximation $\widehat{u}({\bf x})$ of the corresponding exact solution $u({\bf x})$ to the PDE at the point ${\bf x}$. The discriminator aims to identify whether the pair $({\bf x}, \widehat{u}({\bf x}))$ is a real exact solution to the PDE. Similar to the classical GANs, the weights of GA-PINNs are determined by using the minimax game training between the generator and the discriminator. When they reach the Nash equilibrium, {\it i.e.,} the discriminator is trained to be incapable of identifying whether $({\bf x}, \widehat{u}({\bf x}))$ is a real exact solution, the generator will approximate the exact solution $u({\bf x})$ accurately. However, since the labeled samples are insufficient, the generator outputs could be far away from the exact solutions to the PDE, even though the discriminator has be trained to be capable of identifying the fake samples. To overcome this shortcoming, at each training iteration, we further minimize the PI loss ${\rm L}_{\rm PINN}$ (or the weighted PI loss ${\rm L}^{\rm PW}_{\rm PINN}$ given in \eqref{eq:pw.pinnloss}) to fine-tune the generator weights after the minimax game training. The experimental results support the effectiveness of the proposed GA-PINNs and show that GA-PINNs outperform 
PINNs in many well-known PDEs.


\subsection{Related Works}\label{sec:related}

The universal approximation theory, given by \citet{hornik1989multilayer}, provides a theoretical guarantee for using the neural networks to numerically solve PDEs. \citet{lagaris1998artificial} expressed the solution to a PDE as a summation of two parts. The first part can be directly obtained to satisfy the boundary conditions in some simple situations. The second part is the output of a neural network that is trained by only minimizing the equation loss ${\rm L}_f$. In PINNs, the boundary conditions are also encoded into the network and the network weights are obtained by minimizing the PI loss ${\rm L}_{\rm PINN}$ \citep{dissanayake1994neural,raissi2019physics}.

\citet{yu2017deep} proposed the deep ritz method for using neural networks to solve the variational problem 
\begin{equation*}
	\min _{u \in H}  I(u):=\int_{\Omega}\left(\frac{1 }{2}|\nabla u(x)|^{2}-f(x) u(x)\right) \mathrm{d} x, 
\end{equation*}

where the integral $I(u)$ will be discretized by selecting some points from $\Omega$. Since 
some PDEs can be converted into the equivalent variational forms, the deep ritz method can also be used to solve these PDEs. However, it is still challenging to treat some specific boundary conditions of PDEs. 

As addressed by \citet{dissanayake1994neural,raissi2019physics}, the weights of PINNs are obtained by using the batch gradient descent (BDG) method, where the weights are updated by taking the derivative of the loss function computed on all samples. Instead, \citet{sirignano2018dgm} proposed the deep Galerkin method (DGM) that adopts the stochastic gradient descent (SGD) to update the weights of PINNs. In each iteration, the objective function is computed on a mini-batch of points taken from $\Omega$ and $\partial_i \Omega$ randomly. Taking advantage of the randomness brought from the SGD method, DGM could be more suitable to numerically solving the high-dimensional PDEs than the classical PINNs. In this paper, PINNs are still referred to the PINNs with the BDG method if no confusion arises. 

The rest of this paper is organized as follows. In Section \ref{sec:sw}, we introduce the PW method to improve the training performance of PINNs, In Section \ref{sec:ga-pinn}, we show the structure of GA-PINNs and the minimax game training. The numerical experiments are arranged in Section \ref{sec:experiment} and the last section concludes the paper.

\section{Point-weighting Method}\label{sec:sw}

As shown in Tab. \ref{tab:threeholes}, the main difficulty of training PINNs lies in the sharp fluctuation appearing in the minimization of the equation loss ${\rm L}_{f}$ and the boundary-condition losses ${\rm L}_{b_i}$. To improve the training performance of PINNs, we introduce a point-weighting (PW) method to speed up the process of minimizing ${\rm L}_{f}$ and ${\rm L}_{b_i}$. 

Consider the $i$-th boundary condition $\mathcal{B}_i[u]({\bf x})= g_i ({\bf x})$ (${\bf x} \in \partial_i \Omega$). Let $\{ {\bf x}_i^{(m)} \}_{m=1}^{M_i}$ be a set of points taken from the boundary $\partial_i \Omega$, and let $e_i>0$ be the desired level of the boundary-condition error. Define a binary surrogate function
\begin{equation*}
	\beta({\bf x}^{(m)}_{i};e_{i})
	= \begin{cases}
		+1, & \big\|\widehat{u}\big({\bf x}^{(m)}_{i}\big)-u\big({\bf x}^{(m)}_{i}\big)\big\|^{2} \leq  e_{i}; \\
		-1, & \big\|\widehat{u}\big({\bf x}^{(m)}_{i}\big)-u\big({\bf x}^{(m)}_{i}\big)\big\|^{2}>e_{i},
	\end{cases} 
\end{equation*}
which signifies whether the point ${\bf x}^{(m)}_{i}$ satisfies the accuracy requirement of the $i$-th boundary condition. For convenience of presentation, the points leading to $\beta({\bf x}^{(m)}_{i};e_i) = 1$ are called easy-to-learn (EL) points, and otherwise called hard-to-learn (HL) points

The PW method splits the process of minimizing the boundary-condition loss $ {\rm L}_{b_i}$ into two stages: at the first stage, the weights of EL points are increased to stabilize the network as quickly as possible; and at the second stage, the weights of HL points are increased to make the network fit these HL points and then to improve the network performance.

Accordingly, the $i$-th weighted boundary-condition loss at the $k$-th ($k\in \mathbb{N}$) iteration is defined as:  
\begin{equation}\label{eq:boundary.pw}
	{\rm L}^{\rm PW}_{b_i,k}=   \sum_{m=1}^{M_{i}} \omega^{(m)}_{i,k}  \big\|\mathcal{B}_i[\widehat{u}]\big({\bf x}^{(m)}_{i}\big) - g_i \big({\bf x}^{(m)}_{i}\big)\big\|^{2}, 
\end{equation}

where $\omega^{(m)}_{i,k}$ is the point weight of ${\bf x}^{(m)}_{i}$ at the $k$-th iteration with $\sum_{m=1}^{M_i} \omega^{(m)}_{i,k} =1$. When $k=0$, it is clear that the initial weights are $\omega^{(m)}_{i,0} = \frac{1}{M_{i}}$.


Borrowing the idea from the weighting strategy of the Adaboost method \citep{freund1997decision}, after each iteration of training the network, the point weight $\omega^{(m)}_{i,k}$ is updated in the following way:
\begin{equation} \label{eq:updateb.pw}
	\left\{\begin{array}{l}
		\rho_{i,k}:=\sum\limits_{ \{m: \beta({\bf x}_i^{(m)};e_i)=-1\}}   \omega^{(m)}_{i,k};\\ \alpha_{i,k}:=q_i \log \frac{1-\rho_{i,k}}{\rho_{i,k}}; \\
		\omega^{(m)}_{i,k+1}:= \frac{ \omega^{(m)}_{i,k} \cdot \exp (-\alpha_{i,k}\cdot \beta({\bf x}_i^{(m)};e_i))}{\sum_{m=1}^{M_i} \omega^{(m)}_{i,k} \cdot \exp (-\alpha_{i,k}\cdot \beta({\bf x}^{(m)}_i;e_i))},
	\end{array}
	\right.  
\end{equation}
where $\rho_{i,k}$ is the summation of the HL-point weights; the hyperparameter $q$ controls the magnitude of updating the weight $\omega^{(m)}_{i,k}$. It is obvious that $\rho_{i,k} = 0.5$ is the borderline between the aforementioned two stages. Moreover, we also introduce an extra hyperparameter $\epsilon$ for the terminal condition: when $ 1 - \rho_{i,k} \leq \epsilon$ is achieved, we deem that there has been no HL point left and then terminate the training.

Similarly, given a point set $\{ {\bf x}^{(n)} \}_{n=1}^{N} \subset \Omega$, define the weighted equation loss at the $k$-th iteration ($k\in \mathbb{N}$) as 
\begin{equation}\label{eq:equation.pw}
	{\rm L}^{\rm PW}_{f,k}=  \sum_{n=1}^{N} \omega^{(n)}_{k}  \big[\mathcal{L}[\widehat{u}]\big({\bf x}^{(n)}\big)-f\big({\bf x}^{(n)}\big)\big]^{2} 
\end{equation}
with $\sum_{n=1}^N \omega^{(n)}_{k} =1$. Then, given a desired level of equation error $e$, the point weight $\omega^{(n)}_{k}$ of ${\bf x}^{(n)}$ is updated as follows:  
\begin{equation}\label{eq:updatef.pw}
	\left\{\begin{array}{l}
		\rho_{k}:=\sum\limits_{ \{n: \beta({\bf x}^{(n)};e)=-1\}}   \omega^{(n)}_{k};\\
		\alpha_{k}:=q \log \frac{1-\rho_{k}}{\rho_{k}}; \\
		\omega^{(n)}_{k+1}:= \frac{ \omega^{(n)}_{k} \cdot \exp (-\alpha_{k}\cdot \beta({\bf x}^{(n)};e))}{\sum_{n=1}^N \omega^{(n)}_{k} \cdot \exp (-\alpha_{k}\cdot \beta({\bf x}^{(n)};e))}.
	\end{array}
	\right. 
\end{equation}

Interestingly, as shown in Tab. \ref{tab:pw}, the PW method cannot influence the smooth minimization process. Therefore, the PW method has a good applicability in practice, and can be directly used to train PINNs. Accordingly, the loss function for PINN training with the PW method (called PINN+PW) is expressed as follows:
\begin{equation}\label{eq:pw.pinnloss}
	{\rm L}^{\rm PW}_{\rm PINN} := {\rm L}^{\rm PW}_{f,k} + \lambda \sum_{i=1}^I {\rm L}^{\rm PW}_{b_i,k}. 
\end{equation}

\section{Generative Adversarial Physics-informed Neural Networks}\label{sec:ga-pinn}

Let ${\cal S} := \{{\bf x}^{(n)}\}_{n=1}^N$ and ${\cal S}_i :=\{{\bf x}_i^{(m)}\}_{m=1}^{M_i}$ ($1\leq i\leq I$) be the sets of points taken from the domain and its boundary $\partial_i \Omega$, respectively. We also consider a small size of labeled samples ${\cal S}_T:=\{({\bf x}_T^{(j)}, u_T^{(j)})\}_{j=1}^J$, where ${\bf x}_T^{(j)}$ are the points taken from the domain $\Omega$ and $u_T^{(j)}$ are the exact solutions to the PDE at ${\bf x}_T^{(j)}$ with $J \ll N, M$. 

As stated above, since the labeled samples are insufficient, the minimization of \eqref{eq:MSE2} not only fails to refine PINNs but also influences the training performance provided by the minimization of ${\rm L}_{\rm PINN}$. To overcome this shortcoming, we integrate the GA mechanism with the structure of PINNs to form the proposed GA-PINNs. Moreover, in view of the difficulty of training GA-PINNs (it is still challenging to efficiently train individual PINNs or GANs in the literature), we also adopt the PW method to improve the training performance of GA-PINNs.

\begin{figure*}[htbp]
	
	\begin{center} 
		\centerline{\includegraphics[width=1\columnwidth]{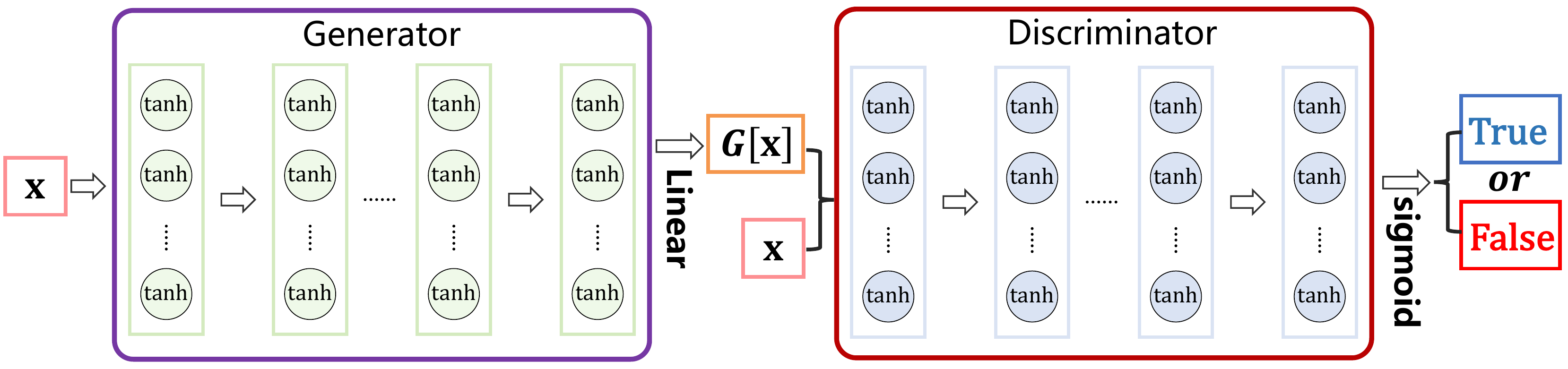}}
		
		\caption{The structure of GA-PINNs. }
		
		\label{fig:ga-pinn}
	\end{center}
\end{figure*}

As shown in Fig. \ref{fig:ga-pinn}, the generator of GA-PINNs inherits the structure of PINNs, where all hidden nodes are activated by using the tanh($\cdot$) functions and the output nodes are activated by using the linear function $y=x$. Its input ${\bf x}$ is a point in the domain $\Omega$ and its output $G[{\bf x}]$ is the approximation of the exact solution $u({\bf x})$ to the PDE. The input of the discriminator is either the pair $({\bf x}_T^{(j)}, G[{\bf x}_T^{(j)}] )$ or the labeled sample $({\bf x}_T^{(j)}, u_T^{(j)})$. The discriminator output $D[({\bf x},G[{\bf x}])]$, activated by using the sigmoid function, is set as a 0-1 node to identify whether the input pair $({\bf x}_T^{(j)}, G[{\bf x}_T^{(j)}])$ is an exact solution to the PDE (labeled as ``1") or not (labeled as ``0"). 

The following are the generative loss and the discriminative loss without the logarithmic operation, respectively:
\begin{equation*}
	\begin{array}{l}
		{\rm L}_{D}=\frac{1}{J} \sum\limits_{j=1}^{J}\big(1-D[({\bf x}_T^{(j)}, u_T^{(j)})]\big)+D [ (  x_{L}^{(j)}, G[x_{L}^{(j)}] )],\\
		{\rm L}_{G}= {\rm L}_T+\frac{1}{J} \sum\limits_{j=1}^{J}\big( 1 -D [(x_{T}^{(j)}, G[x_{T}^{(j)}])] \big).
	\end{array} 
\end{equation*}
The generator weights ${\bf W}_G$ and the discriminator weights ${\bf W}_D$ are updated by minimizing ${\rm L}_{G}$ and ${\rm L}_{D}$, respectively. Since the labeled samples $({\bf x}_T^{(j)}, u_T^{(j)})$ are insufficient, even if the discriminator is trained to be capable of identifying that the pair $(x_{T}^{(j)}, G[x_{T}^{(j)}])$ is a fake labeled sample, the generator could not provide satisfactory approximations of the exact solutions to the PDE. Therefore, at each training iteration, we will minimize ${\rm L}_{\rm PINN}$ (or ${\rm L}^{\rm PW}_{\rm PINN}$) to fine-tune the generator weights ${\bf W}_G$ so as to correct the generator outputs after the minimax game between the generator and the discriminator.

Moreover, since the discriminator is responsible to guide the behavior of generator, we first update the discriminator weights and then update the generator weights in order. Another important thing is the ratio of the learning rates $\eta_G$ and $\eta_D$ for minimizing the generative loss ${\rm L}_{G}$ and the discriminator loss ${\rm L}_{D}$, respectively. Here, we set $\eta_{G}: \eta_{D}=1: 5$ to raise the degree of training the discriminator so as to maintain a powerful guidance to the generator. The workflow of training GA-PINN is given in Alg. \ref{alg:ga-pinn}, and it also illustrates the workflow of training GA-PINN with PW method (called GA-PINN+PW) by replacing ${\rm L}^{\rm PW}_{\rm PINN}$ with ${\rm L}_{\rm PINN}$ in Line 5:
\begin{algorithm}[htbp] 
	\renewcommand{\algorithmicrequire}{\textbf{Input:}}
	\renewcommand{\algorithmicensure}{\textbf{Output:}}
	\caption{The process of training GA-PINN (resp. GA-PINN+PW)}\label{alg:ga-pinn}
	\begin{algorithmic}[1]	
		\REQUIRE $\{{\bf x}^{(n)}\}_{n=1}^N$, $\{{\bf x}_{i}^{(m)}\}_{m=1}^{M_{i}}$, $\{({\bf x}_{T}^{(j)},u_T^{(j)})\}_{j=1}^J$, $e$, $e_i$, $\eta_D$, $\eta_G$, $K$, $q$, $q_i$, $\epsilon$; 
		\ENSURE ${\bf W}_{G}$, ${\bf W}_{D}$;
		\STATE Initialize ${\bf W}^{(0)}_{G}$, ${\bf W}^{(0)}_{D}$; let $\omega^{(m)}_{i,0} = \frac{1}{M_{i}}$ and $\omega^{(n)}_{0} = \frac{1}{N}$;
		\FORALL{ $k=1,2,\cdots,K$; }
		\STATE Update ${\bf W}_{\rm D}$ by minimizing the discriminative loss ${\rm L}_{ D}$: ${\bf W}_{D}^{(k+1)} = {\bf W}_{\rm D}^{(k)} -  \eta_D\cdot\frac{\partial  {\rm L}_{\rm D}}{\partial {\bf W}_{ D}}$; 
		\STATE Update ${\bf W}_{\rm G}$ by minimizing the generative loss ${\rm L}_{\rm G}$: ${\bf W}_{G}^{(k,1)} = {\bf W}_{G}^{(k)} -  \eta_G\cdot\frac{\partial  {\rm L}_{G}}{\partial {\bf W}_{G}}$; 
		\STATE Update ${\bf W}_G^{(k,1)}$ by minimizing ${\rm L}_{\rm PINN}$: ${\bf W}_{G}^{(k+1)} = {\bf W}_G^{(k,1)} - \eta_G\cdot \frac{\partial  {\rm L}_{\rm PINN}}{\partial {\bf W}_{G}}$ (resp. by minimizing ${\rm L}^{\rm PW}_{\rm PINN}$: ${\bf W}_{G}^{(k+1)} = {\bf W}_G^{(k,1)} - \eta_G\cdot \frac{\partial  {\rm L}^{\rm PW}_{\rm PINN}}{\partial {\bf W}_{G}}$).
		\ENDFOR
	\end{algorithmic}
	\label{Alg:GA-PINN}
\end{algorithm}

\begin{table*}[t!]
	\centering 
	\caption{Examining the effect of the PW method for different testing PDEs}
	\label{tab:threeholes}
	\resizebox{\linewidth}{!}{
		\begin{threeparttable}
			\begin{tabular}{  c | c | c | c | c }
				\hline
				& Minimizing ${\rm L}_{f}$ & Minimizing  ${\rm L}_{b}$ & Training Error (${\rm L}_{\rm PINN}$) & Testing Error ({\rm NRMSE})\\ 
				\hline
				\makecell[c]{\rotatebox{90}{Burgers}}&\begin{minipage}[b]{0.6\columnwidth}
					\centering
					\raisebox{-.5\height}{
						\includegraphics[width=\linewidth]{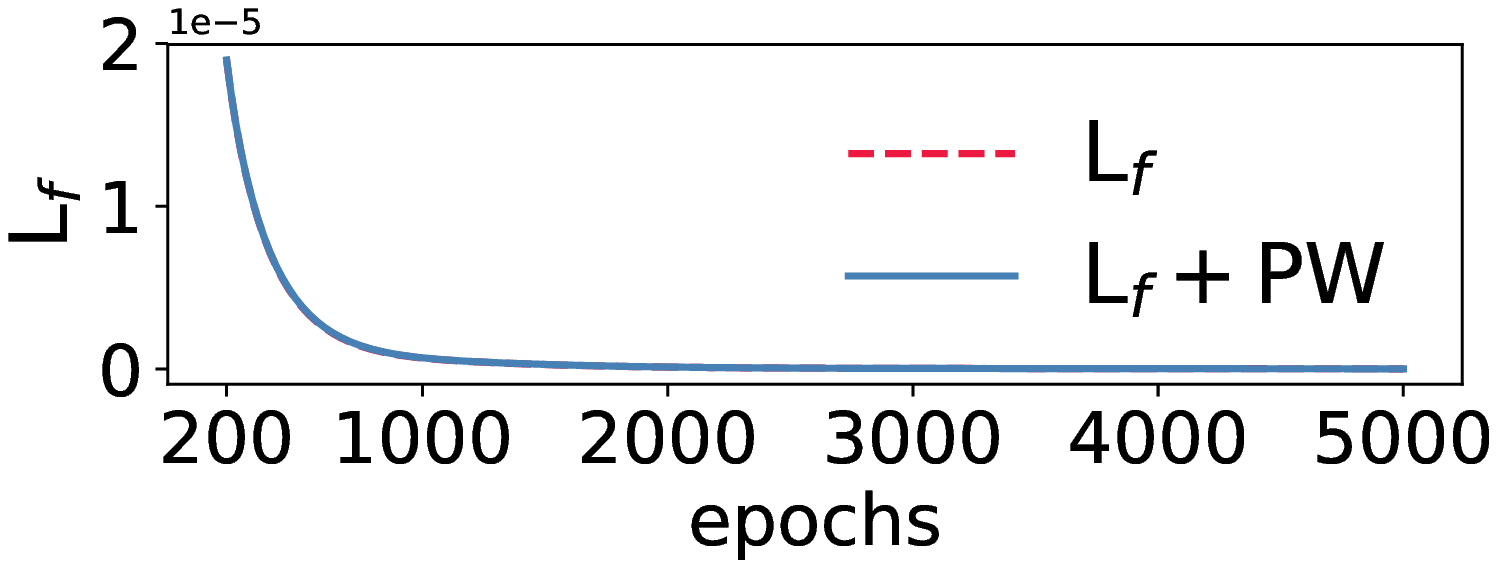}}
				\end{minipage}
				&\begin{minipage}[b]{0.6\columnwidth}
					\centering
					\raisebox{-.5\height}{
						\includegraphics[width=\linewidth]{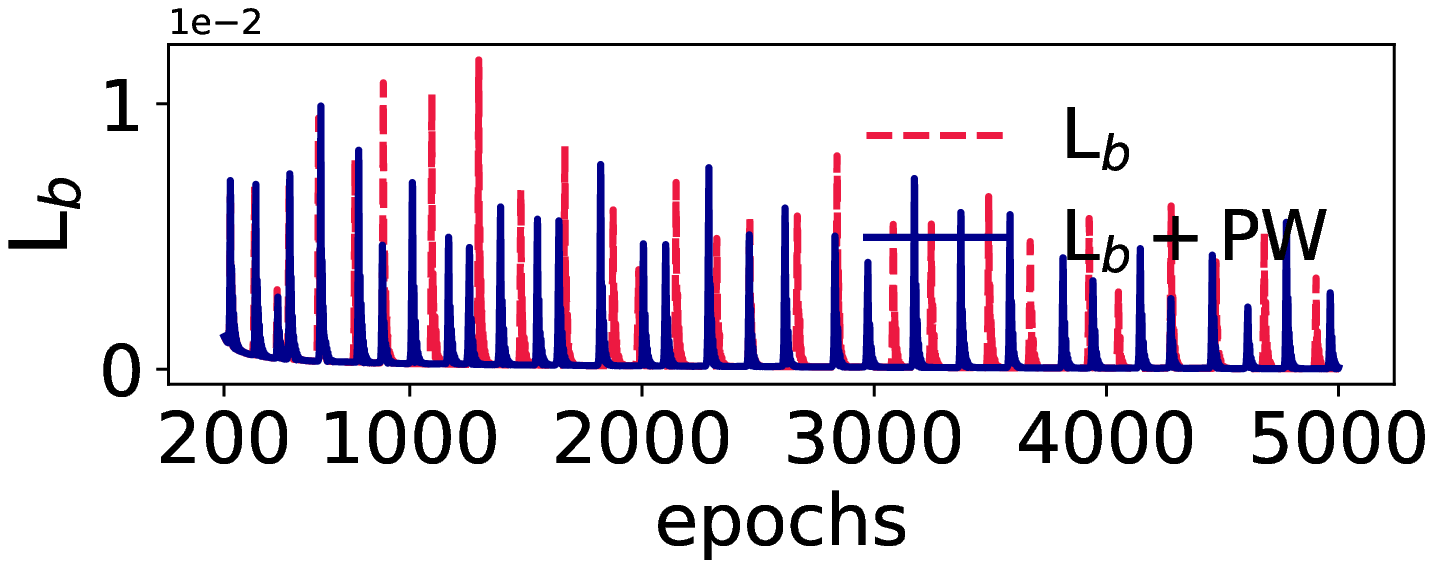}}
				\end{minipage}
				&\begin{minipage}[b]{0.6\columnwidth}
					\centering
					\raisebox{-.5\height}{
						\includegraphics[width=\linewidth]{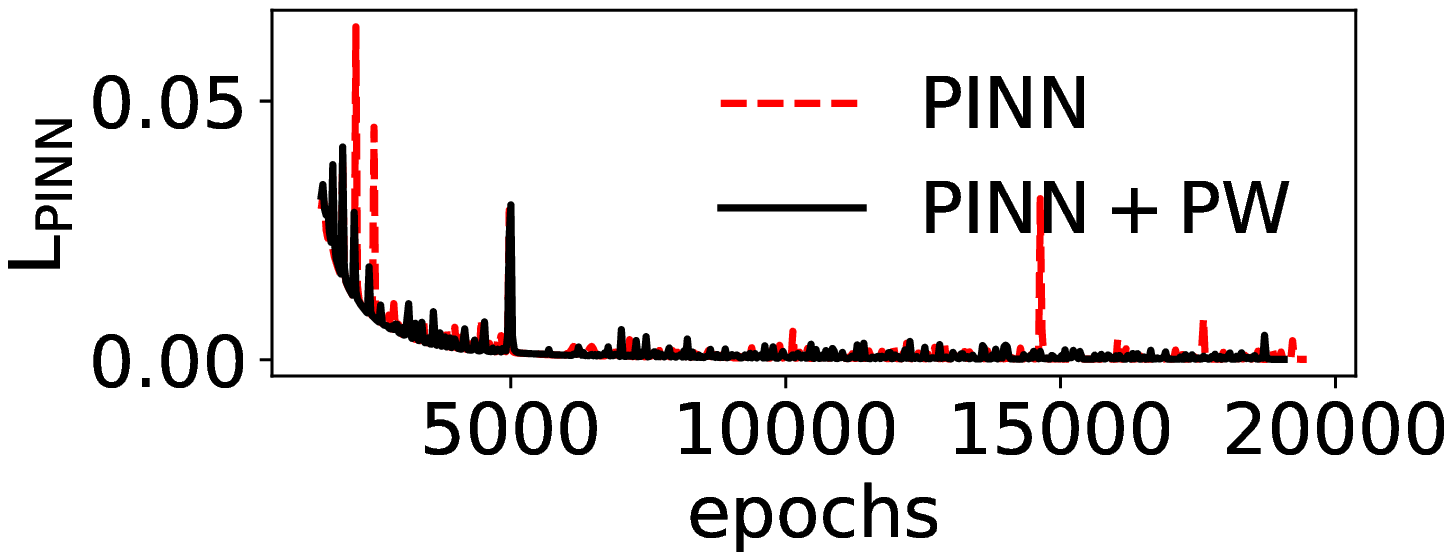}}
				\end{minipage}
				&\begin{minipage}[b]{0.6\columnwidth}
					\centering
					\raisebox{-.5\height}{
						\includegraphics[width=\linewidth]{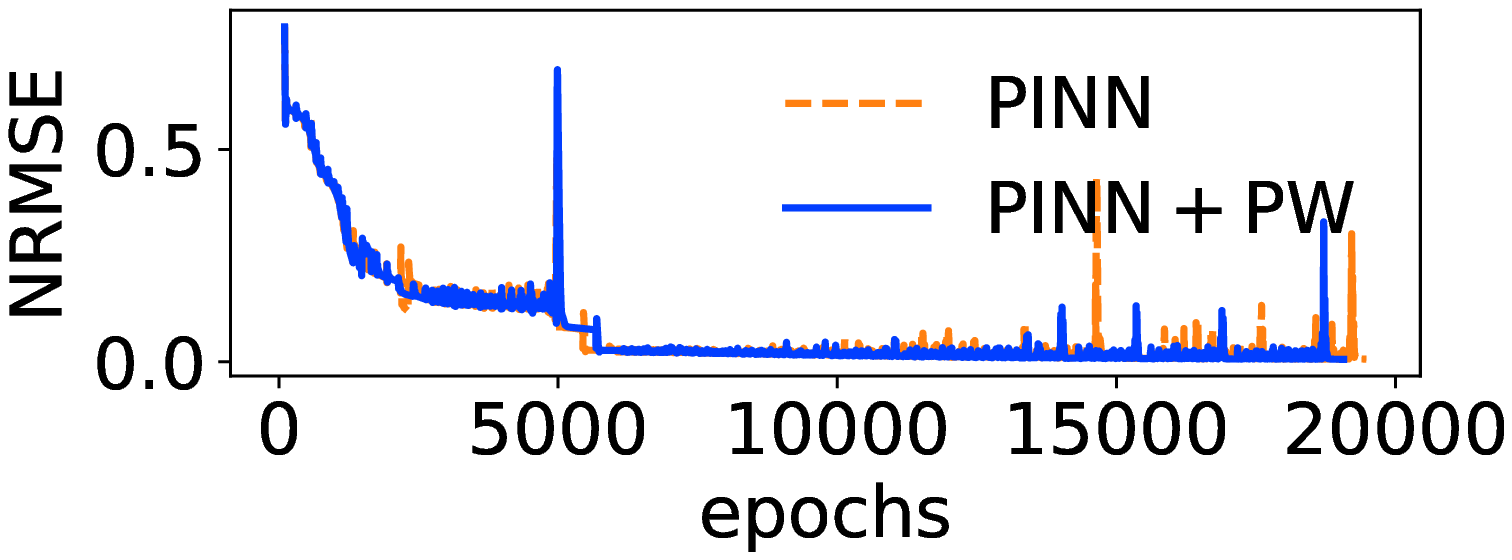}}
				\end{minipage}
				\\ 
				\hline
				\makecell[c]{\rotatebox{90}{Schrodinger}}&\begin{minipage}[b]{0.6\columnwidth}
					\centering
					\raisebox{-.5\height}{
						\includegraphics[width=\linewidth]{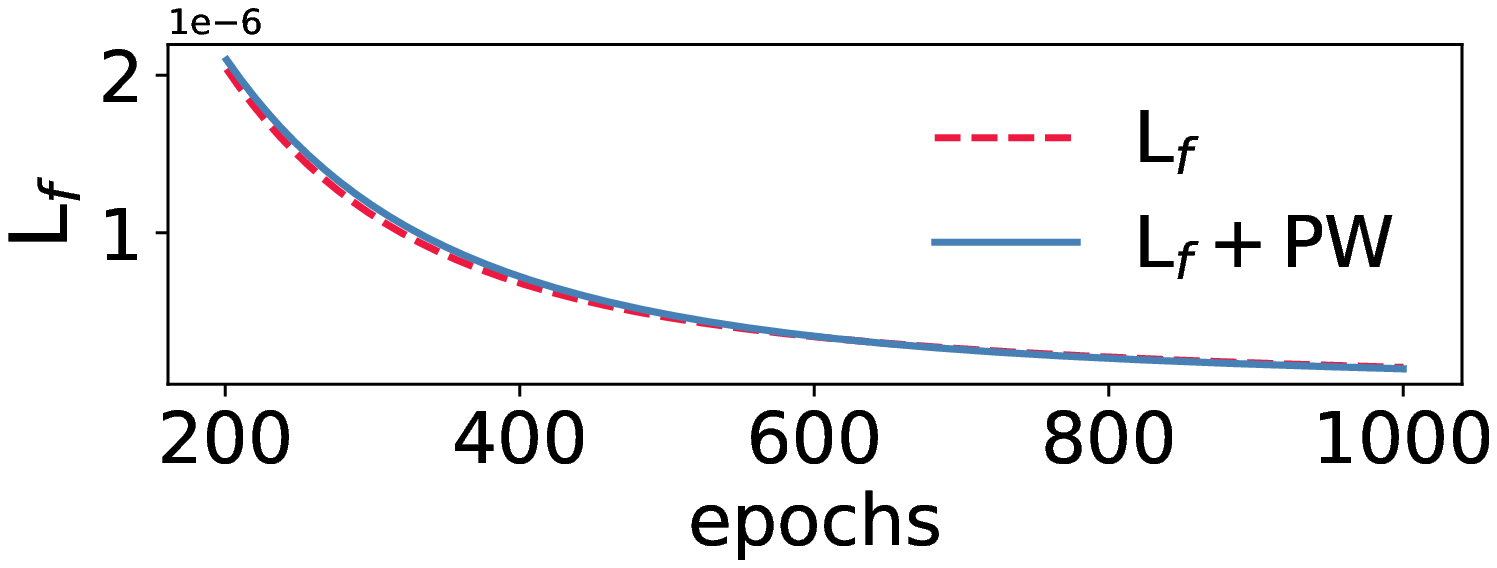}}
				\end{minipage}
				&\begin{minipage}[b]{0.6\columnwidth}
					\centering
					\raisebox{-.5\height}{
						\includegraphics[width=\linewidth]{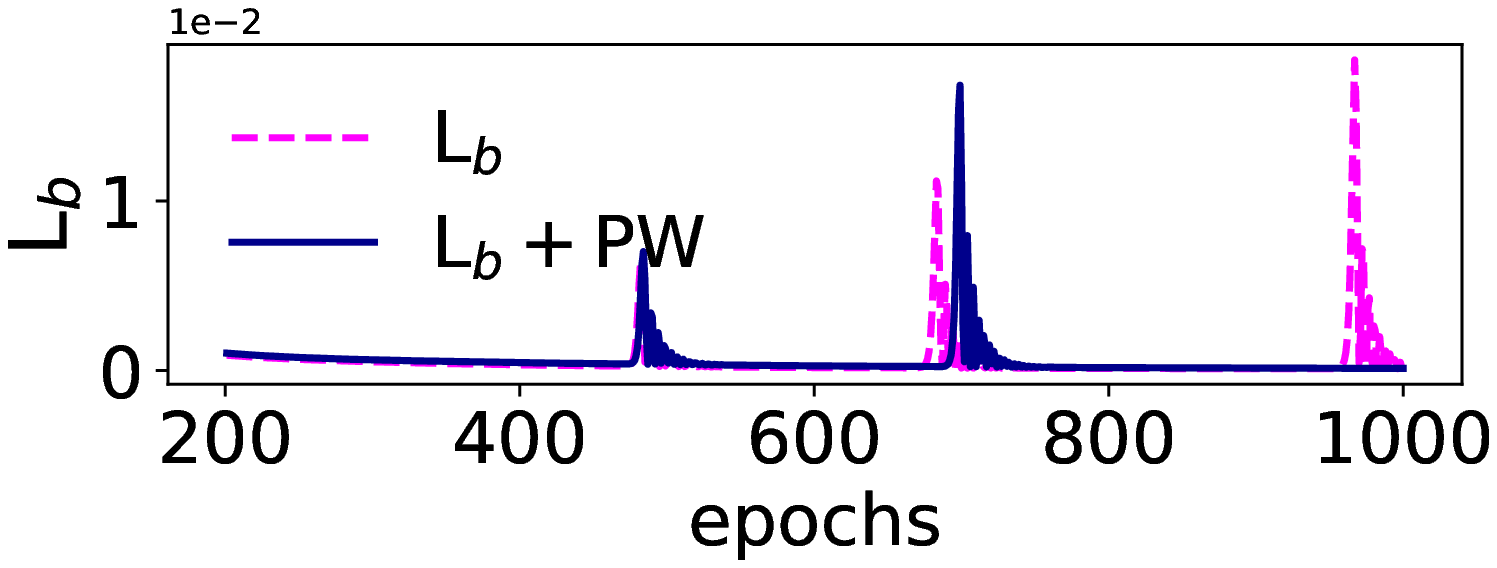}}
				\end{minipage}
				&\begin{minipage}[b]{0.6\columnwidth}
					\centering
					\raisebox{-.5\height}{
						\includegraphics[width=\linewidth]{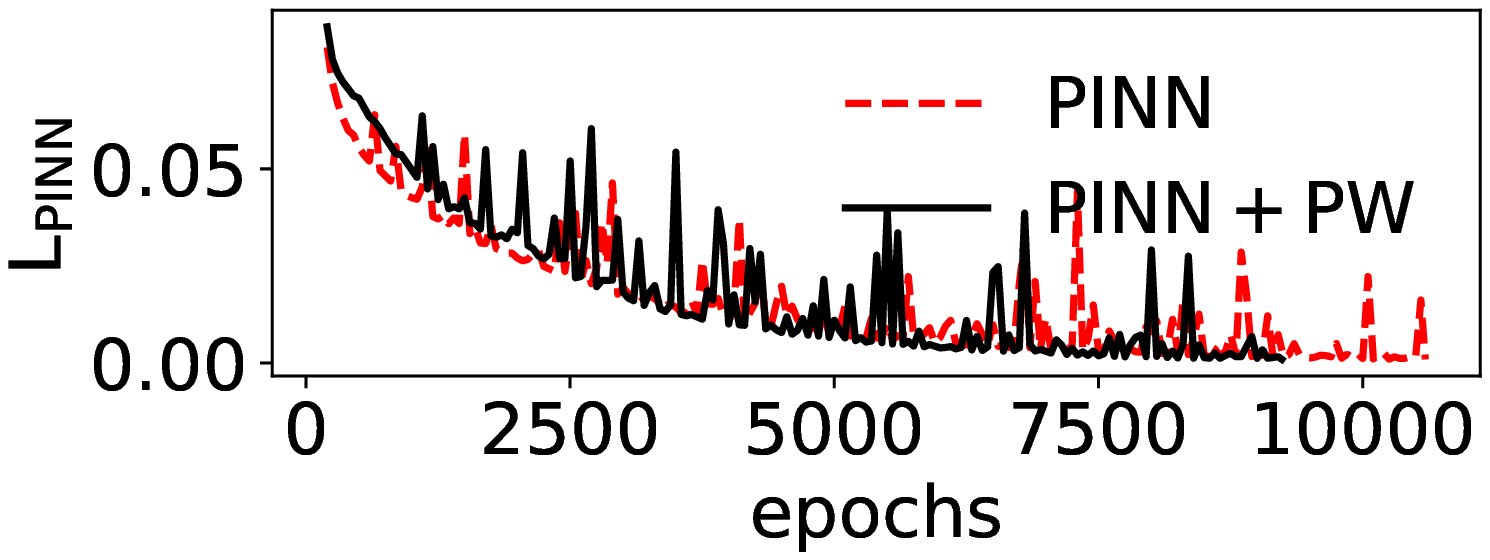}}
				\end{minipage}
				&\begin{minipage}[b]{0.6\columnwidth}
					\centering
					\raisebox{-.5\height}{
						\includegraphics[width=\linewidth]{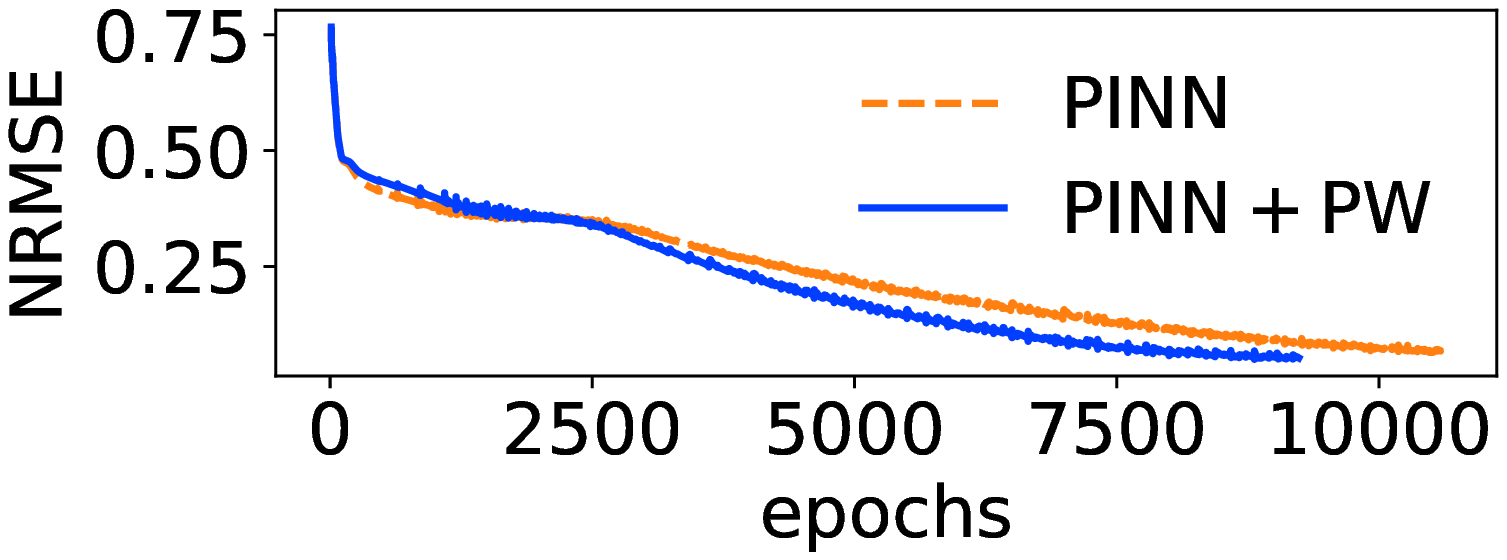}}
				\end{minipage}
				\\ 
				\hline
				\makecell[c]{\rotatebox{90}{Helmholtz}}&\begin{minipage}[b]{0.6\columnwidth}
					\centering
					\raisebox{-.5\height}{
						\includegraphics[width=\linewidth]{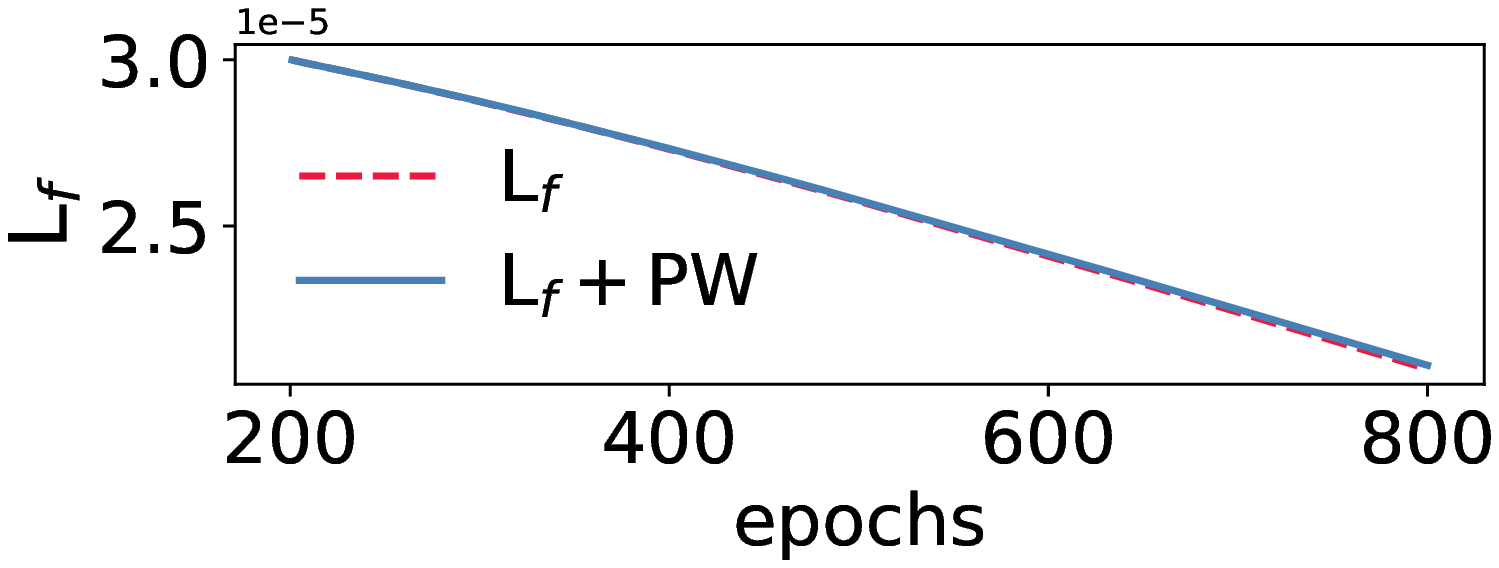}}
				\end{minipage}
				&\begin{minipage}[b]{0.6\columnwidth}
					\centering
					\raisebox{-.5\height}{
						\includegraphics[width=\linewidth]{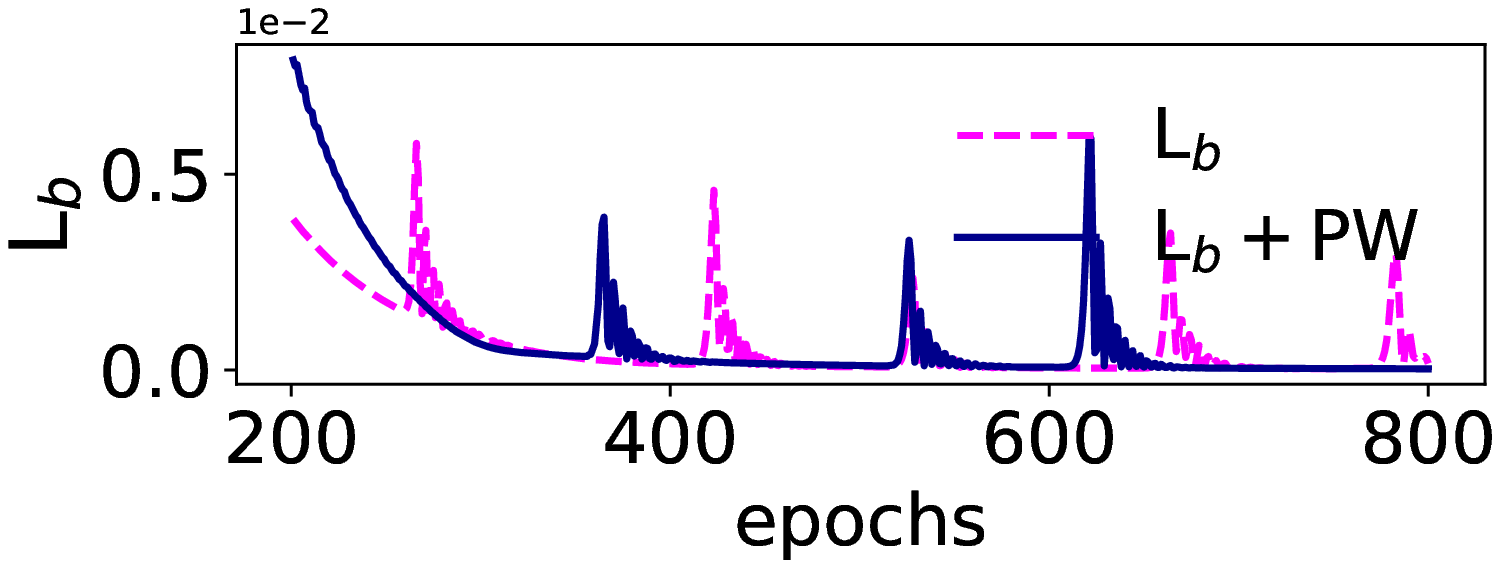}}
				\end{minipage}
				& \begin{minipage}[b]{0.6\columnwidth}
					\centering
					\raisebox{-.5\height}{
						\includegraphics[width=\linewidth]{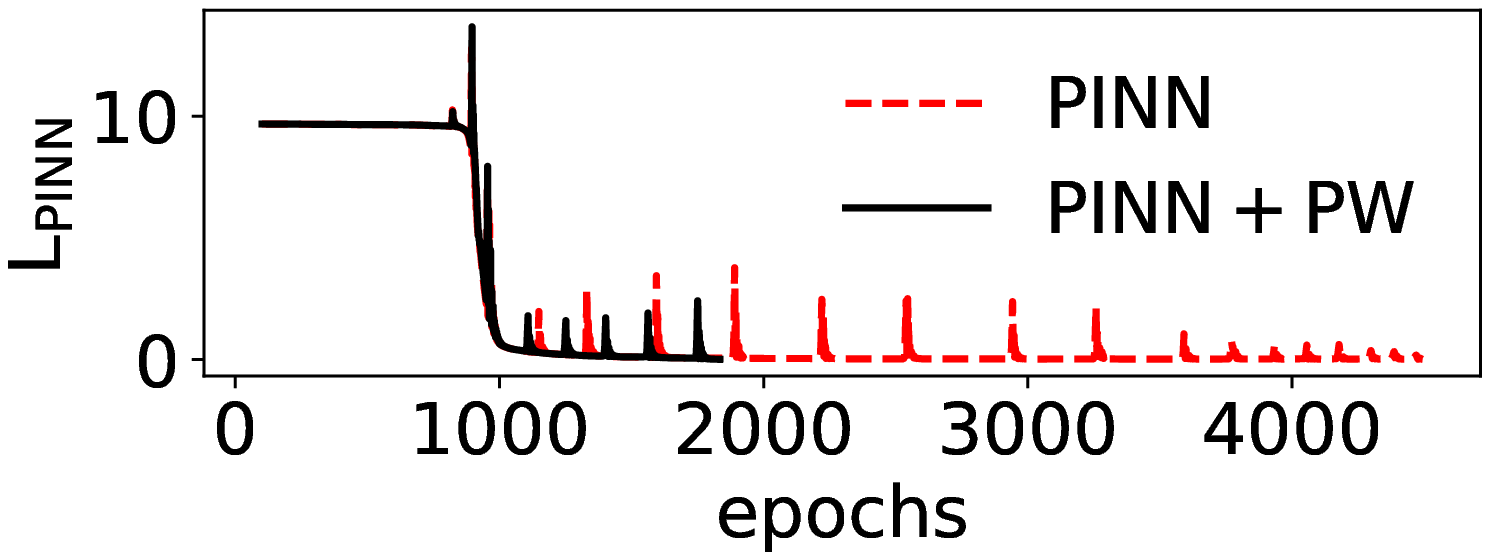}}
				\end{minipage}
				&\begin{minipage}[b]{0.6\columnwidth}
					\centering
					\raisebox{-.5\height}{
						\includegraphics[width=\linewidth]{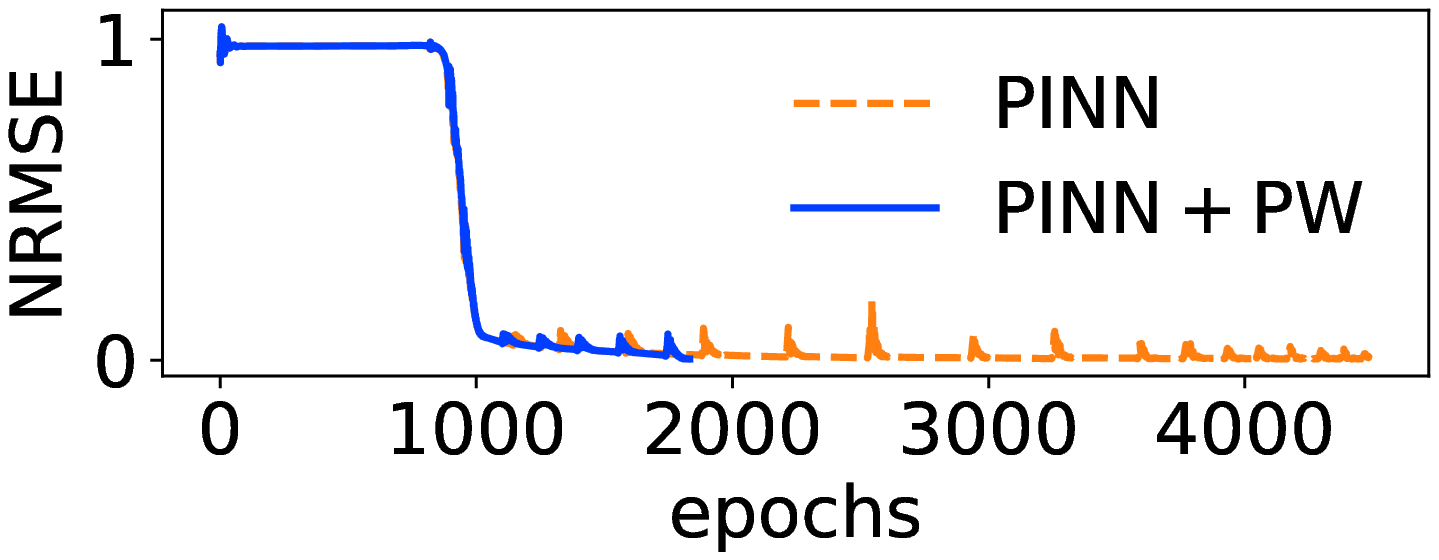}}
				\end{minipage}
				\\ 
				\hline
				\makecell[c]{\rotatebox{90}{Poisson}}&\begin{minipage}[b]{0.6\columnwidth}
					\centering
					\raisebox{-.5\height}{
						\includegraphics[width=\linewidth]{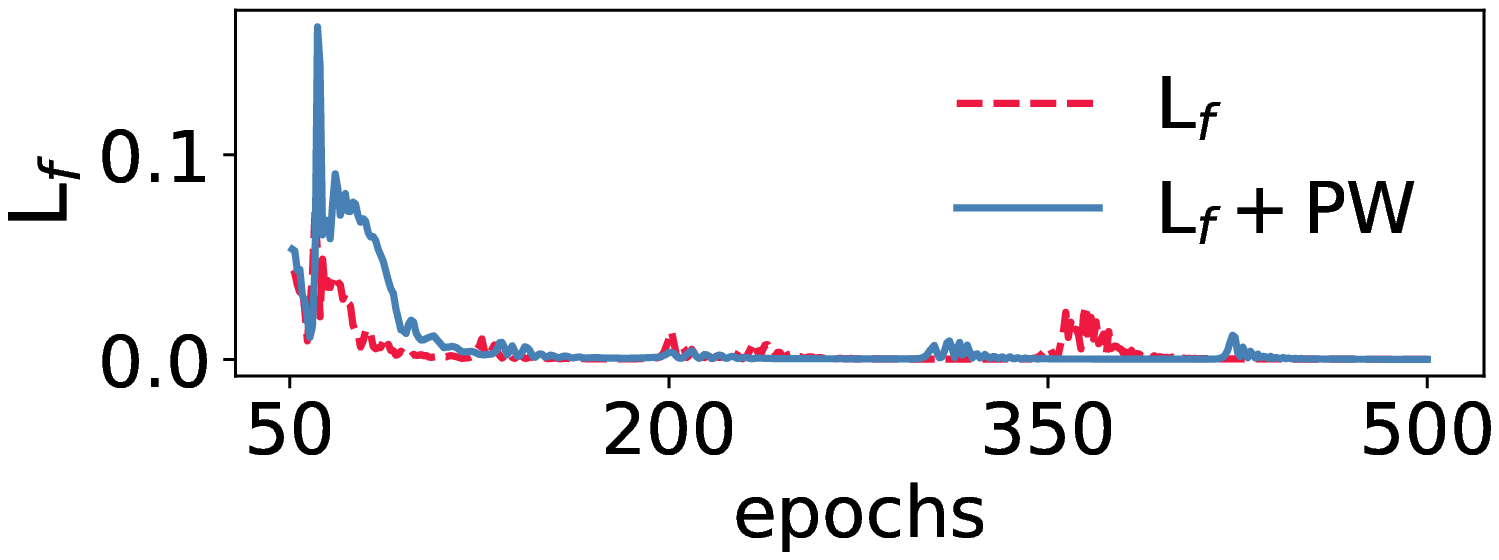}}
				\end{minipage}
				&\begin{minipage}[b]{0.6\columnwidth}
					\centering
					\raisebox{-.5\height}{
						\includegraphics[width=\linewidth]{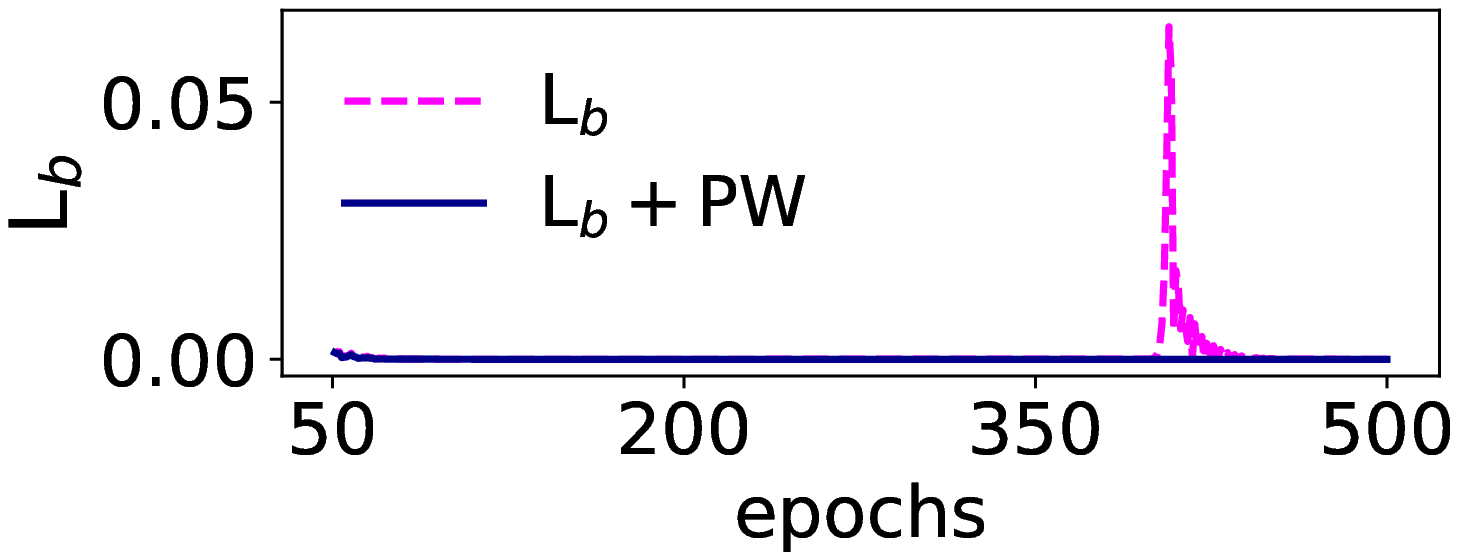}}
				\end{minipage}
				&\begin{minipage}[b]{0.6\columnwidth}
					\centering
					\raisebox{-.5\height}{
						\includegraphics[width=\linewidth]{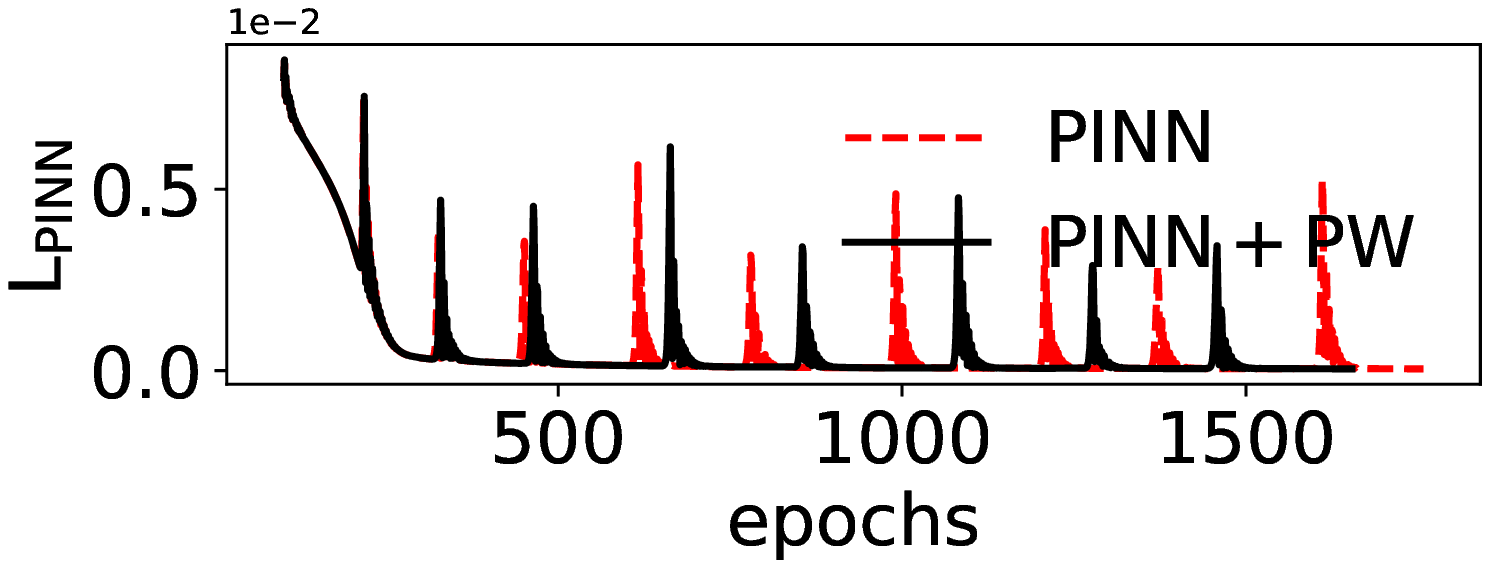}}
				\end{minipage}
				&\begin{minipage}[b]{0.6\columnwidth}
					\centering
					\raisebox{-.5\height}{
						\includegraphics[width=\linewidth]{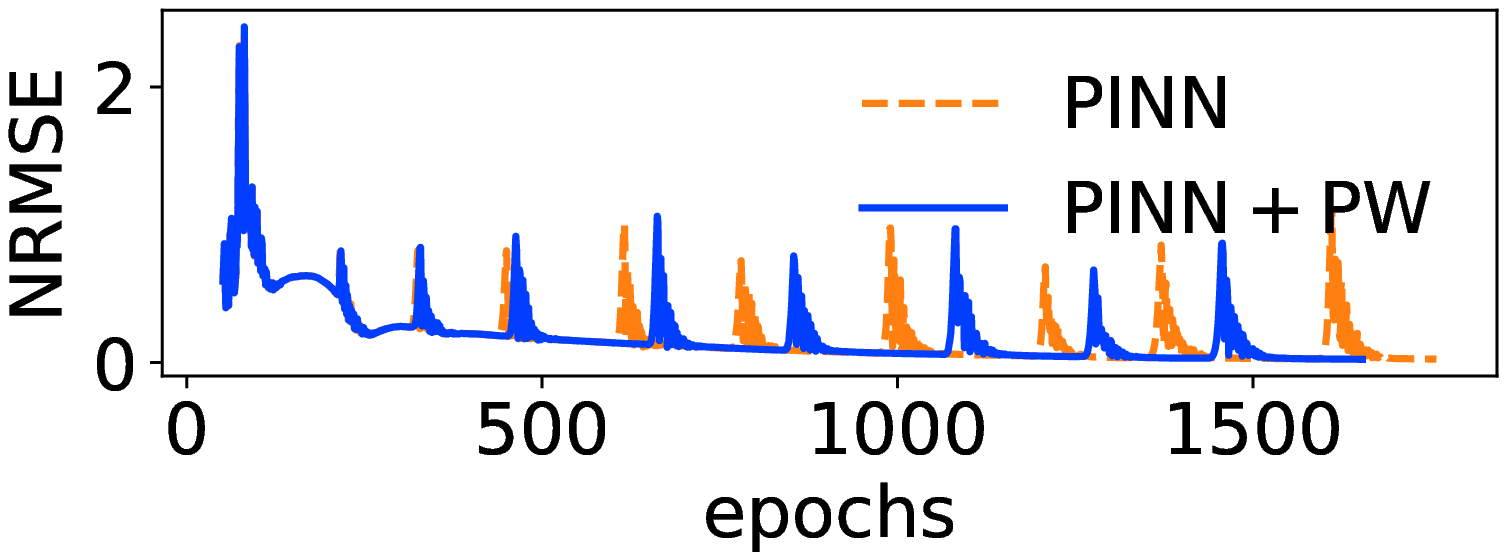}}
				\end{minipage}
				\\ 
				\hline
				\makecell[c]{\rotatebox{90}{HD Poisson}}&\begin{minipage}[b]{0.6\columnwidth}
					\centering
					\raisebox{-.5\height}{
						\includegraphics[width=\linewidth]{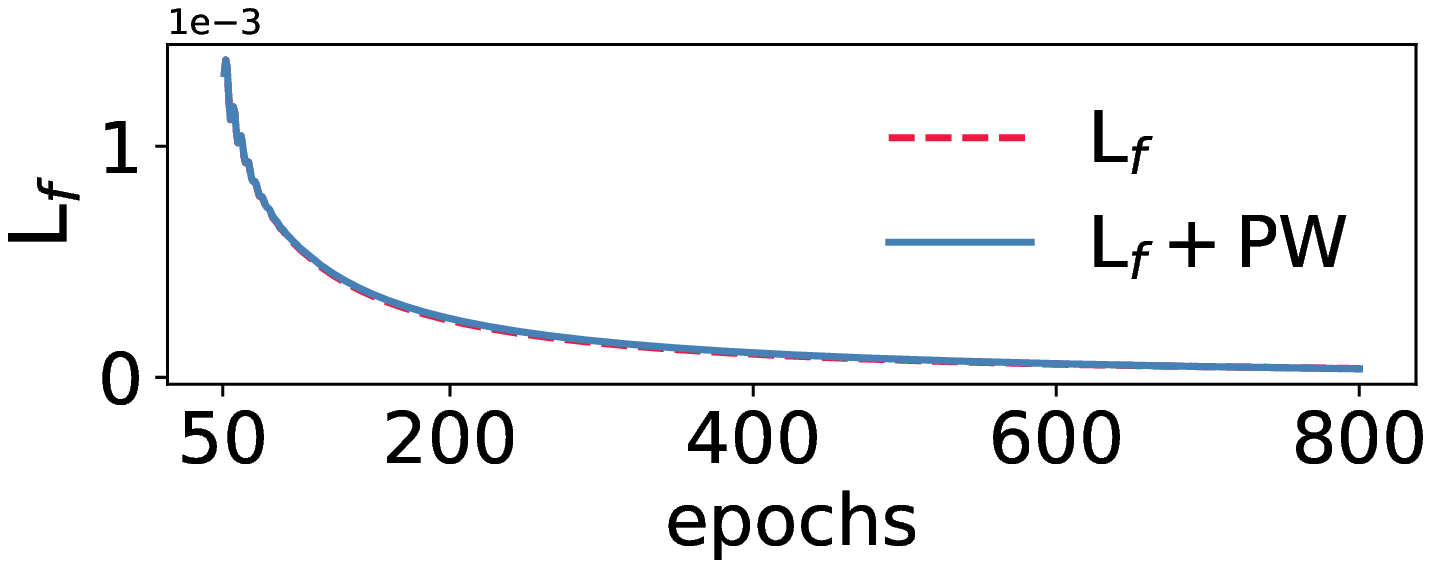}}
				\end{minipage}
				&\begin{minipage}[b]{0.6\columnwidth}
					\centering
					\raisebox{-.5\height}{
						\includegraphics[width=\linewidth]{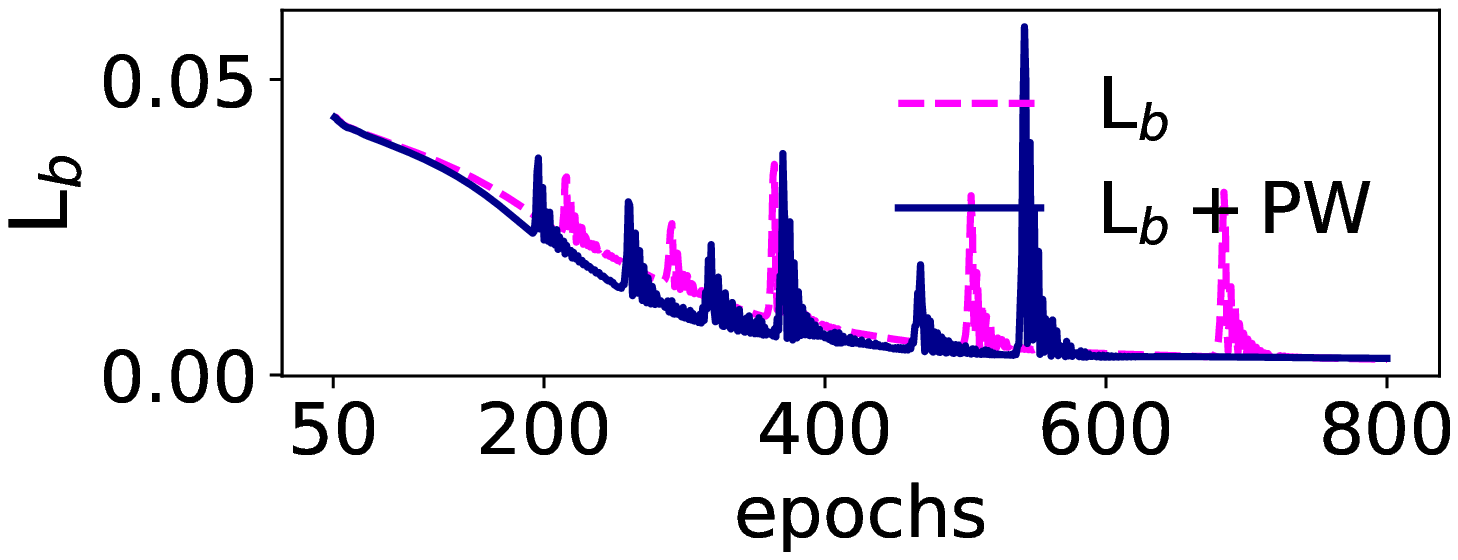}}
				\end{minipage}
				&\begin{minipage}[b]{0.6\columnwidth}
					\centering
					\raisebox{-.5\height}{
						\includegraphics[width=\linewidth]{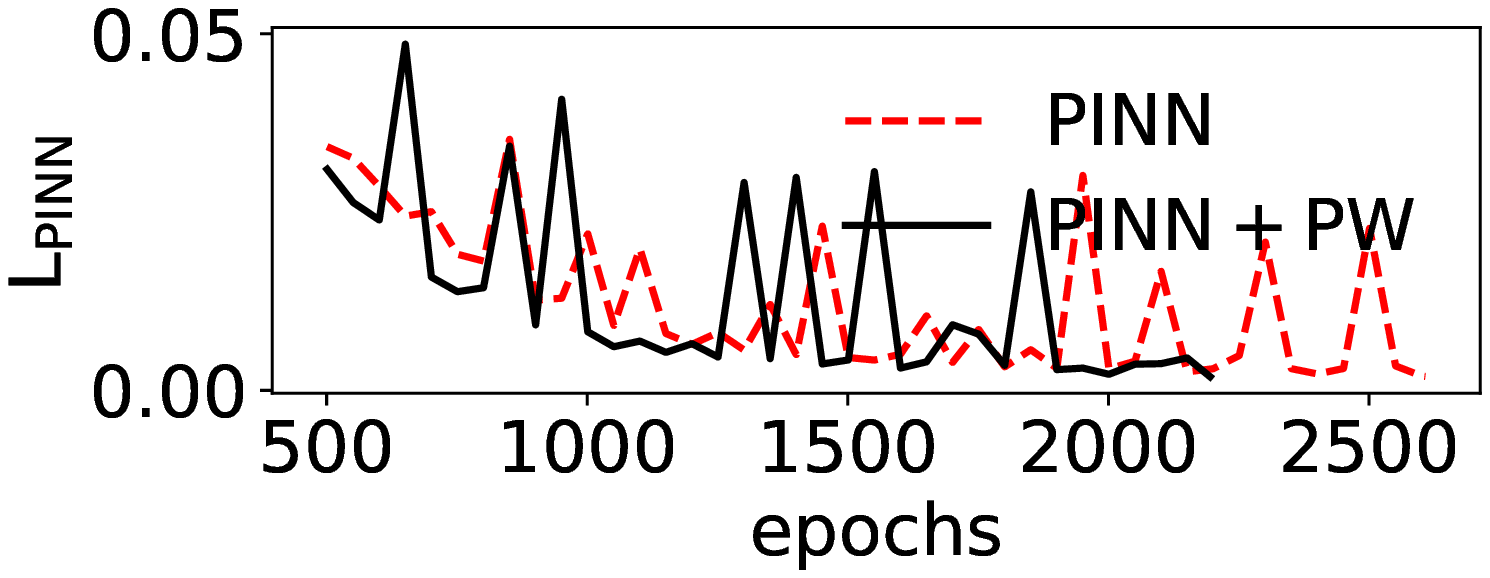}}
				\end{minipage}
				&\begin{minipage}[b]{0.6\columnwidth}
					\centering
					\raisebox{-.5\height}{
						\includegraphics[width=\linewidth]{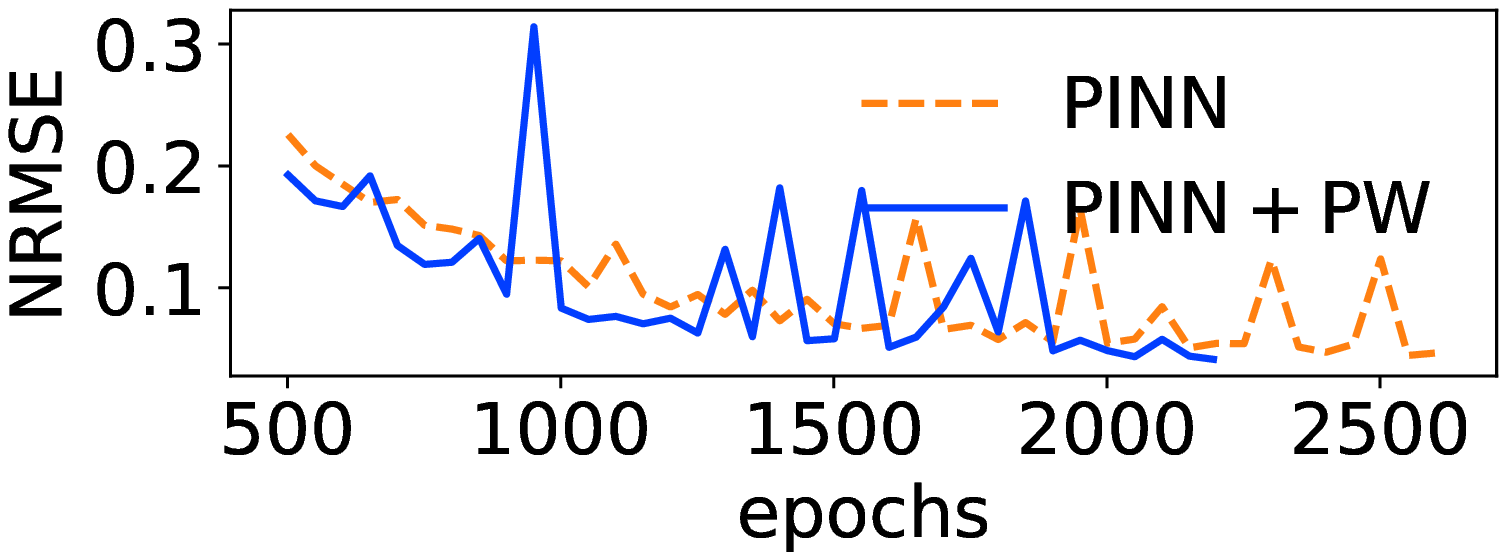}}
				\end{minipage}
				\\ 
				\hline
				\makecell[c]{\rotatebox{90}{Heat}}&\begin{minipage}[b]{0.6\columnwidth}
					\centering
					\raisebox{-.5\height}{
						\includegraphics[width=\linewidth]{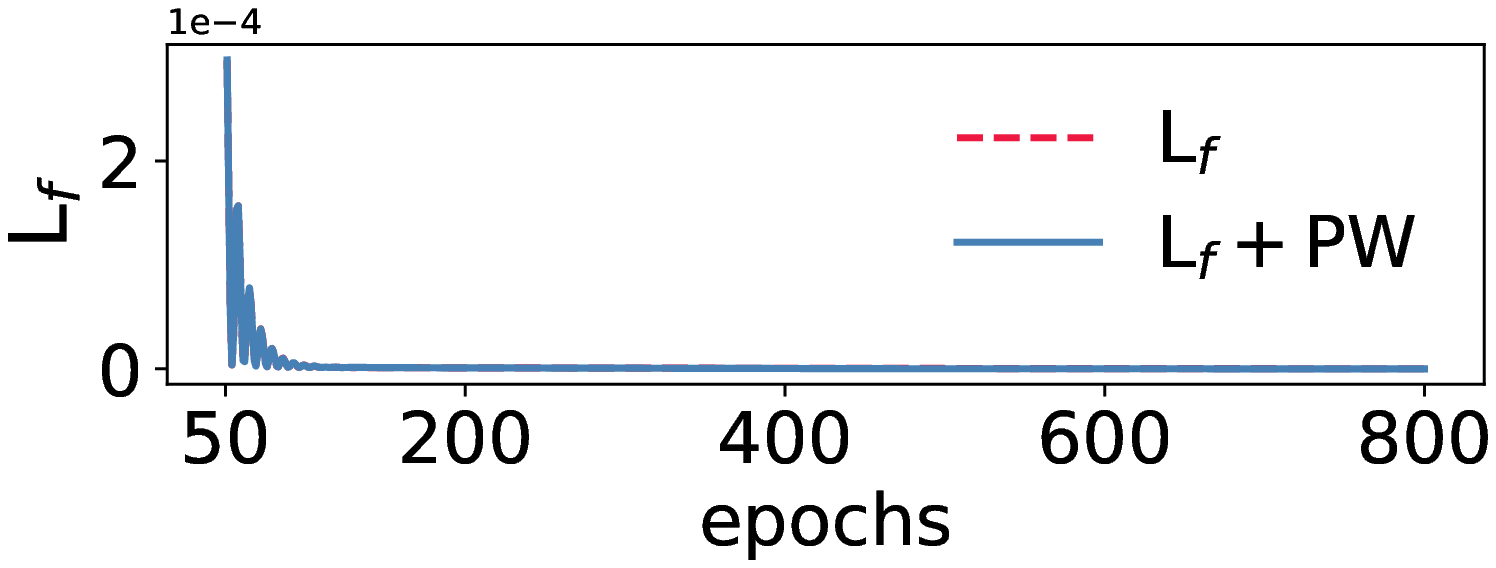}}
				\end{minipage}
				&\begin{minipage}[b]{0.6\columnwidth}
					\centering
					\raisebox{-.5\height}{
						\includegraphics[width=\linewidth]{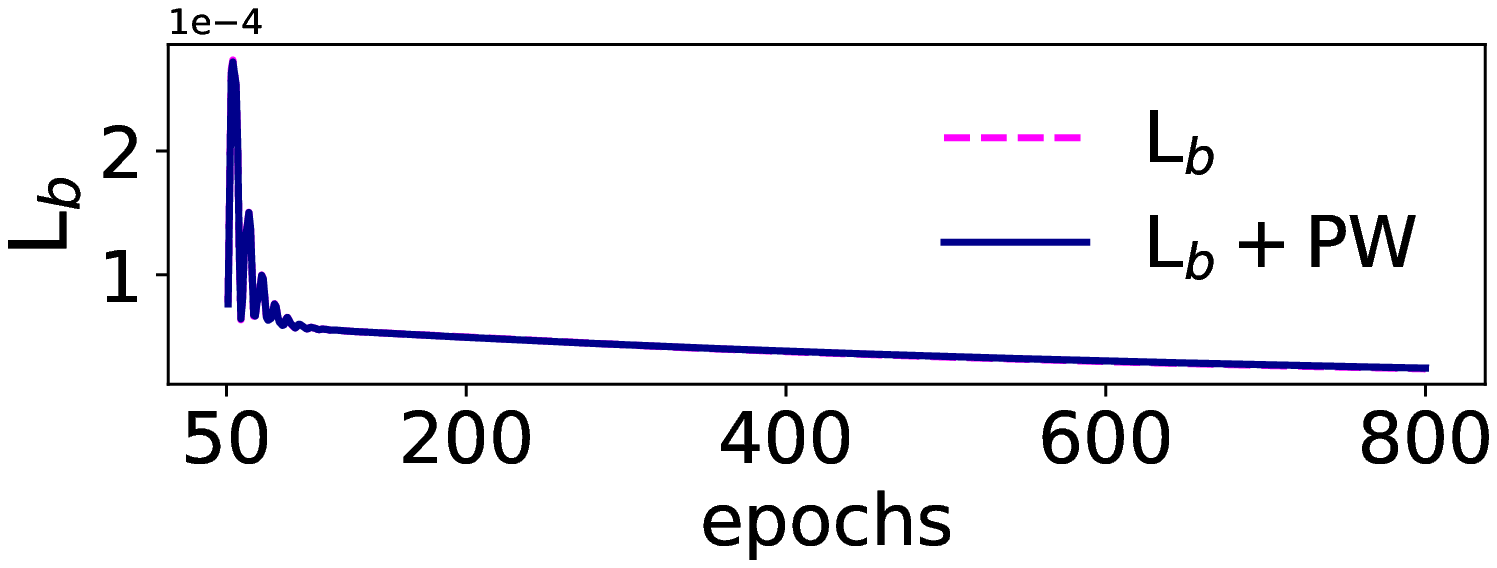}}
				\end{minipage}
				&\begin{minipage}[b]{0.6\columnwidth}
					\centering
					\raisebox{-.5\height}{
						\includegraphics[width=\linewidth]{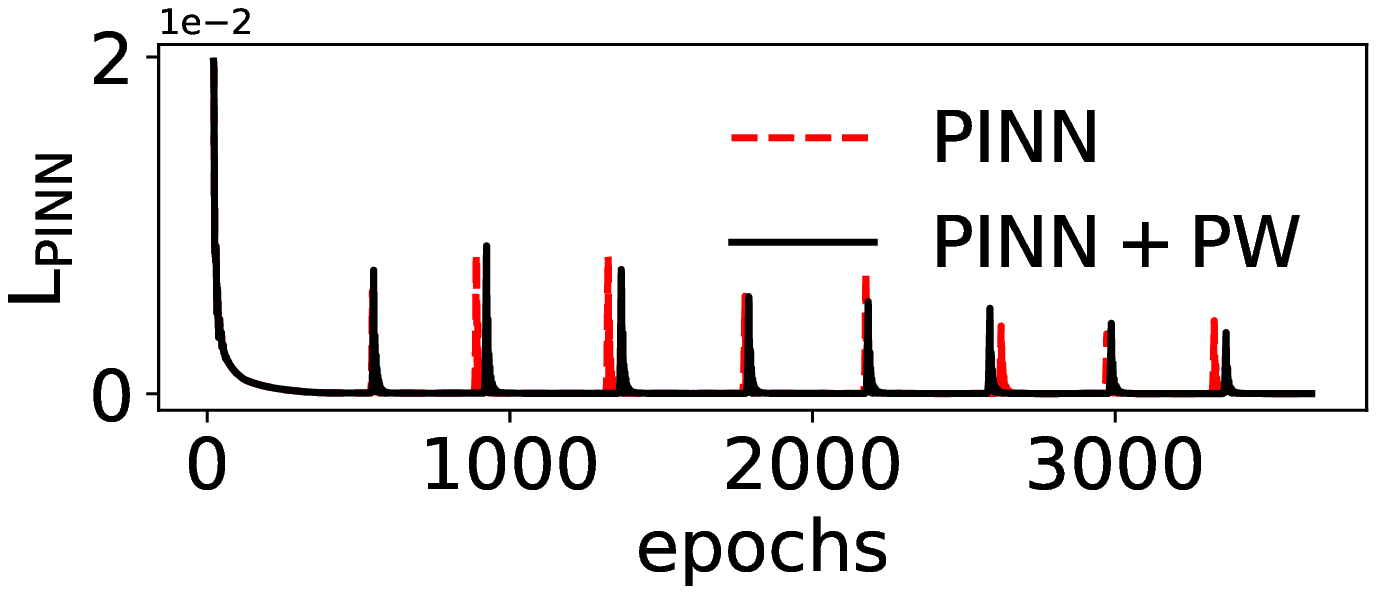}}
				\end{minipage}
				&\begin{minipage}[b]{0.6\columnwidth}
					\centering
					\raisebox{-.5\height}{
						\includegraphics[width=\linewidth]{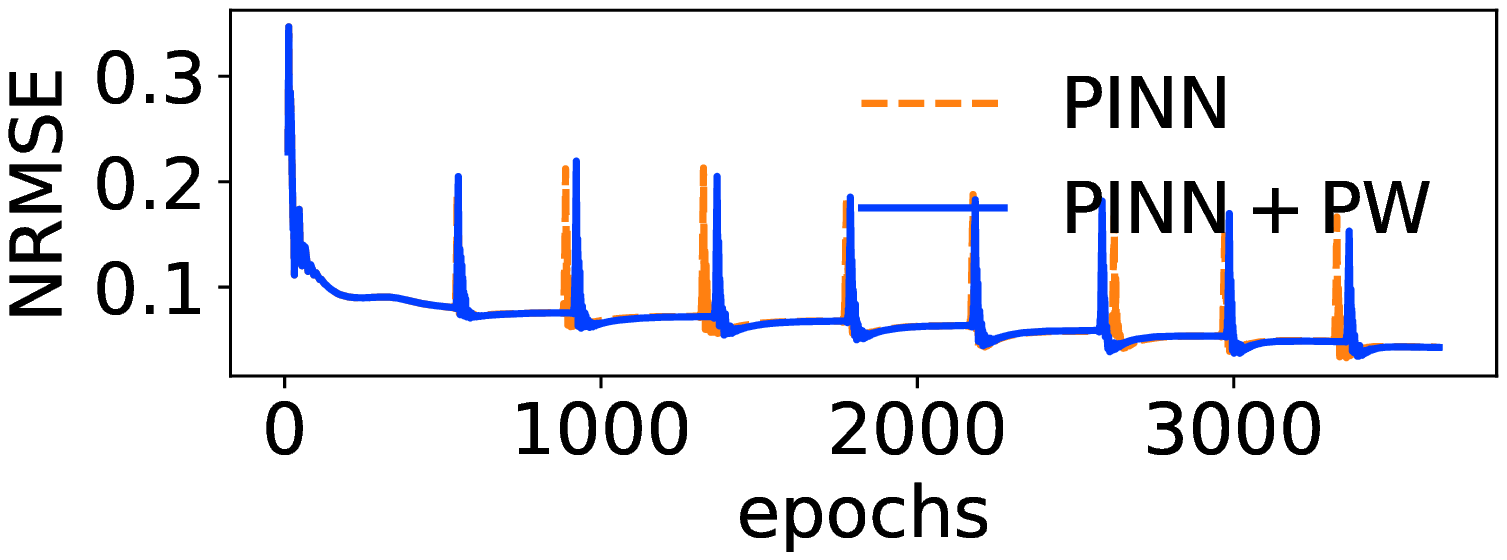}}
				\end{minipage}
				\\ 
				\hline
			\end{tabular}
			\begin{tablenotes}
				\item[*] The first column shows the processes of individually minimizing ${\rm L}_f$. The second column shows the processes of individually minimizing ${\rm L}_b$. The third column shows the curves of the training errors measured by ${\rm L}_{\rm PINN}$ computed on the training samples. The fourth column shows the curves of the testing errors measured by ${\rm NRMSE}$ computed on the testing samples.
			\end{tablenotes}
		\end{threeparttable}\label{tab:pw}
	}
\end{table*}



\section{Numerical Experiments}\label{sec:experiment}

In this section, we conduct the numerical experiments to verify the effectiveness of the proposed PW method and GA-PINNs for numerically solve PDEs in the situation that a small size of labeled samples are available. We mainly concern with the following issues: 1) whether the PW method can speed up the process of training a PINN and improve its performance; and 2) whether the proposed GA-PINNs outperform PINNs in such a situation.

\begin{table*}[htbp]
	\centering
	\caption{Six testing PDE problems}
	\label{tab:pde}
	\resizebox{0.9\linewidth}{!}{
		\begin{threeparttable}
			\begin{tabular}{  l | l | l  }
				\hline
				PDE & Main Equation & Boundary Condition \\
				\hline
				\makecell[l]{Burgers} & $u_{t}+u u_{x}-(0.01 / \pi) u_{x x}=0$,\;\;$(x,t)\in  [-1,1] \times [0,1]$ & $
				\left\{\begin{array}{l}
					u(0, x)=-\sin (\pi x);\\
					u(t,-1)=u(t, 1)=0.
				\end{array}\right.$\\
				\hline
				\makecell[l]{Poisson } & 
				$\Delta u=-\sin (\pi x) \sin (\pi y), \;\; (x, y) \in (0,1) \times(0,1)$
				&$\left\{
				\begin{array}{l l}
					u(x,y) =0, &(x, y) \in \{0,1\}\times(0,1) ; \\
					u(x,y) =0, &(x, y) \in (0,1)\times \{0,1\} .
				\end{array}
				\right.
				$\\
				\hline
				\makecell[l]{Helmholtz} &
				$\Delta u+k^2 u =0$,\;\;$(x, y) \in (0,1) \times(0,1)$ & $\left\{
				\begin{array}{l l}
					u(x,y) =\sin(k x), &(x, y) \in \{0,1\}\times(0,1) ; \\
					u(x,y) =\sin(k x), &(x, y) \in (0,1)\times \{0,1\} .
				\end{array}\right.$\\
				\hline
				\makecell[l]{Schrodinger}&
				$i h_{t}+0.5 h_{x x}+|h|^{2} h=0$,\;\;$ (x,t) \in[-5,5]\times[0, \pi / 2]$ &
				$\left\{\begin{array}{l}
					h(0, x)=2 \operatorname{sech}(x), \\
					h(t,-5)=h(t, 5), \\
					h_{x}(t,-5)=h_{x}(t, 5),
				\end{array}\right.
				$ \\
				\hline
				\makecell[l]{HD Poisson} &
				$-\Delta u=0$,\;\; $\boldsymbol{x} \in(0,1)^{10}$ &
				$\begin{aligned}
					u(\boldsymbol{x})=x_{1}^{2}-x_{2}^{2}+x_{3}^{2}&-x_{4}^{2}+x_{5} x_{6}+x_{7} x_{8} x_{9} x_{10},\\
					& \boldsymbol{x} \in \partial_i(0,1)^{10},\;\; 1\leq i\leq 10;
				\end{aligned}$ \\
				\hline
				\makecell[l]{Heat} &
				$u_{t}-\Delta u=0$,\;\; $(x,y,t)\in  \mathbb{R}^3$ &
				$u(x,y,0)=x-y ,\;\; (x,y)\in \mathbb{R}^2. $\\
				\hline
			\end{tabular}
			\begin{tablenotes}
				\footnotesize
				\item[1] In HD Poisson equation, $ \boldsymbol{x} \in \partial_i(0,1)^{10}$ means that $\boldsymbol{x} \in \underbrace{(0,1)\times(0,1)}_{i-1} \times  \{0,1\} \times \underbrace{(0,1)\times(0,1)}_{10-i} $.
				\item[2] Helmholtz equation, Poisson equation, HD Poisson equation and heat equation have the analytic solutions: $u(x,y) = \sin(kx)$, $u(x,y) = \frac{1}{2 \pi^{2}} \sin (\pi x) \sin (\pi y)$,  $u(\boldsymbol{x})=x_{1}^{2}-x_{2}^{2}+x_{3}^{2}-x_{4}^{2}+x_{5} x_{6}+x_{7} x_{8} x_{9} x_{10}$ and $u(x,y,t) =\frac{1}{(4 \pi t)} \int_{\mathbb{R}^{2}} e^{-\frac{(x-z_{1})^{2}+(y-z_{2})^{2}}{4 t}} (z_1-z_2) dz_{1} dz_{2}  $, respectively. The testing datasets of the three PDEs are obtained by using Latin hypercube sampling on their domains, respectively. 
				\item[3] The testing datasets of Burgers equation and Schrodinger equation are given in the PINN software package \citep{maziarraissi}. 
			\end{tablenotes}
		\end{threeparttable}
	}
\end{table*}

\begin{table*}[htbp]
	\centering 
	\caption{The choices of hyperparameters in the numerical experiments }\label{tab:hyperparameter}
	\resizebox{\linewidth}{!}{
		\begin{threeparttable}
			\begin{tabular}{l l l l l l l l l l l l l l l}
				\hline
				PDE         & $q_{1}$ 			&$e_{1}$					&$q_{2}$ 			& $e_{2}$   			& TC (${\rm L}_{\rm PINN}$)  & $ N $  & $ M $  & $J$ & $ \eta_G $ & $ \eta_P$  & $ \eta_D $  &  ${\rm HL}_G$   &   ${\rm HL}_D$   \\ \hline
				Burgers     & $1\times10^{-4}$  & 0.02 						& $1\times10^{-4}$  & $5\times10^{-4}$ 		& $1\times10^{-4}$ 		& 10000 	& 100	 & 10  & 0.001 	& 0.001  & 0.005 	 & L: 7, N: 20  &  L: 8, N: 20 	\\
				Schrodinger & 0.005   			& $5\times10^{-4}$ 			& 0.005   			& $1\times10^{-4}$ 		& 0.001  				& 20000 	& 100 	 & 10 & 0.001 	& 0.001  & 0.005 	 & L: 4, N: 100  & L: 3, N: 100 	\\
				Helmholtz   & $6\times10^{-5}$ 	& $5\times10^{-4}$			&$6\times10^{-5}$& $5\times10^{-4}$	& $0.01$   				& 20000 	& 200	 & 3  & 0.001 	& $1\times10^{-5}$   & $5\times10^{-5}$ 	 &  L: 4, N: 100  & L: 1, N: 100 	\\
				Poisson     & $5\times10^{-5}$ 	& $5\times10^{-6}$ 			& $5\times10^{-5}$  & $5\times10^{-6}$ 		& $5\times10^{-5}$   	& 5000 		& 100	 & 5  & 0.001 	& $1\times10^{-6}$  & $5\times10^{-6}$  	 &  L: 4, N: 100  & L: 1, N: 100 	\\
				HD Poisson  & 0.001   			& 0.05 		&\makecell[c]{ $q_1=q_2=$\\$ \cdots=q_{10}$}       			& \makecell[c]{ $e_1=e_2=$\\$ \cdots=e_{10}$}				& 0.002  				& 10000 	& 500	 & 100  & 0.001 	& 0.001  & 0.005 	 &  L: 4, N: 100  & L: 1, N: 100 	\\ 
				Heat  & $5\times10^{-5}$  	&  $5\times10^{-6}$  		& $5\times10^{-5}$ 	& $5\times10^{-6}$	& $5\times10^{-6}$ 				& 5000 	& 100	 & 10  & 0.001 	& 0.001  & 0.005 	 &  L: 4, N: 100  & L: 1, N: 100 	\\ \hline
				
			\end{tabular}
			\begin{tablenotes}
				\item[1] $q_1,q_2$ and $e_1,e_2$ are the hyperparameters used in the PW method ({\it cf.} \eqref{eq:updateb.pw}). In HD Poisson, the hyperparameters $q_i$ (resp. $e_i$) ($1\leq i\leq 10$) share the same value $q_1 =\cdots = q_{10} = 0.001$ (resp. $e_1 =\cdots = e_{10} = 0.05$). Especially, we set $q=5\times10^{-5}$ and $e= 0.001$ in \eqref{eq:updatef.pw} for Poisson equation.
				\item[2] TC (${\rm L}_{\rm PINN}$) stands for the termination condition that is the loss ${\rm L}_{\rm PINN}$ computed on the training data reaches the predefined level. 
				\item[3] $N$ and $M$ are the sizes of training data taken from a domain and its boundaries, respectively. Following the suggestions provided by \citet{raissi2019physics}, we adopt a relatively size number $N_b$ to maintain the stability of training process.  Moreover, $J$ is the size of labeled samples. 
				\item[4] $ \eta_G $, $ \eta_D $ and $ \eta_P$ are the learning rates for minimizing the generative loss ${\rm L}_G$, the discriminator loss ${\rm L}_D$ an the loss ${\rm L}_{\rm PINN}$, respectively.
				\item[5] ${\rm HL}_G$ (resp. ${\rm HL}_D$) stands for the hidden layers of the generator (resp. discriminator), where ``L: 7, N: 20" means that there are $7$ hidden layers and each hidden layer has $20$ nodes.
			\end{tablenotes}
		\end{threeparttable}
	} 
\end{table*}

\begin{table*}[htbp]
	\centering 
	\caption{The epoch numbers and testing NRMSEs of the four models when the training process satisfies the termination conditions.}\label{tab:performance}\label{tab:NRMSE}
	\resizebox{\linewidth}{!}{
		\begin{tabular}{c|c|c|c|c|c|c|c|c|c|c|c|c}
			\hline
			\multirow{2}*{\diagbox{Method}{PDE}} &\multicolumn{2}{c|}{Burgers}& \multicolumn{2}{c|}{Schrodinger} & \multicolumn{2}{c|}{Helmholtz}   & \multicolumn{2}{c|}{Poisson}  & \multicolumn{2}{c|}{HD Poisson} & \multicolumn{2}{c}{Heat}\\
			\cline{2-13}
			~ &epoch & NRMSE & epoch & NRMSE & epoch & NRMSE & epoch & NRMSE& epoch & NRMSE & epoch & NRMSE\\
			\hline
			PINN &19475 & $6.39\times10^{-3}$ &10593  &0.0669  &4494  &$4.87\times10^{-3}$ &1758   &0.0242 &2608  &0.0435 &3640   &0.0432\\
			\hline
			PINN+PW &19068  & $5.89\times10^{-3}$ &9235 &0.0523 &1834 &\underline{{\it $3.62\times10^{-3}$}}& {\bf 1654}   &\underline{0.0230} & {\bf 2195} &0.0424  & 3649  & 0.0428\\
			\hline
			GA-PINN &{\bf 17585} & \underline{$4.69\times10^{-3}$} &6080 &\underline{0.0122} &2433  &$4.36\times10^{-3}$  &3704  &0.0255 &4354  &0.0469 &{\bf 1847}   &0.0099\\
			\hline
			GA-PINN+PW &18000 & $4.81\times10^{-3}$ & {\bf 5798} &0.0156 &{\bf 1800} &$4.74\times10^{-3}$  &3475 &\underline{0.0238} &2909 & 0.0399 &2047 &\underline{0.0089}\\
			\hline
			DGM  &50000 & $0.0265$ & 20000 &0.2266 &8000 &$6.28\times10^{-3}$  &6000 &0.1025&6000 &\underline{0.0317} &6000 &0.0181\\
			\hline
	\end{tabular}}
\end{table*}

\begin{table*}[htbp]
	\centering 
	\caption{Heat maps of the discrepancies between the exact solutions and the numerical solutions provided by different models}
	\label{tab:threeholes}
	\resizebox{\linewidth}{!}{
		\begin{threeparttable}
			\begin{tabular}{  c | c | c | c | c | c}
				\hline
				& {\rm PINN} & {\rm PINN+PW} & {\rm GA-PINN} & {\rm GA-PINN+PW} & {\rm DGM}\\ 
				\hline
				\makecell[c]{\rotatebox{90}{{\small Burgers}}}&\begin{minipage}[b]{0.6\columnwidth}
					\centering
					\raisebox{-.5\height}{
						\includegraphics[width=\linewidth]{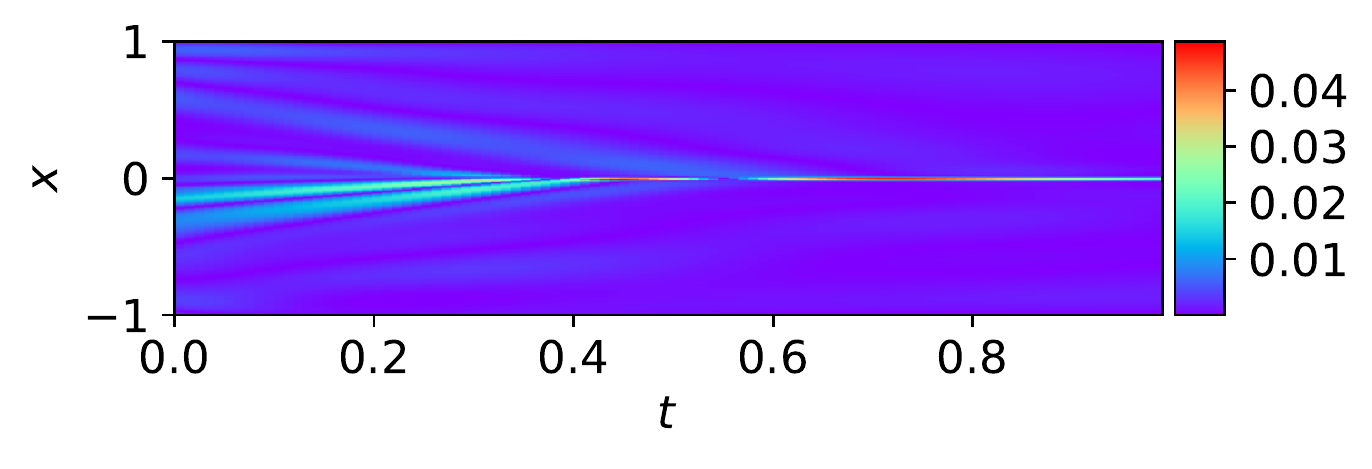}}
				\end{minipage}
				&\begin{minipage}[b]{0.6\columnwidth}
					\centering
					\raisebox{-.5\height}{
						\includegraphics[width=\linewidth]{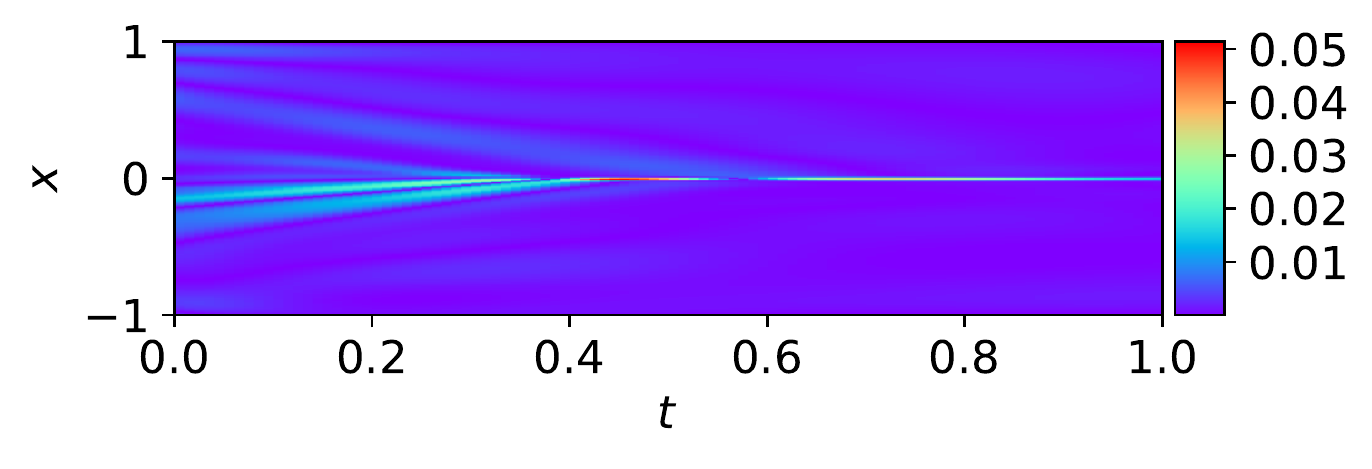}}
				\end{minipage}
				&\begin{minipage}[b]{0.6\columnwidth}
					\centering
					\raisebox{-.5\height}{
						\includegraphics[width=\linewidth]{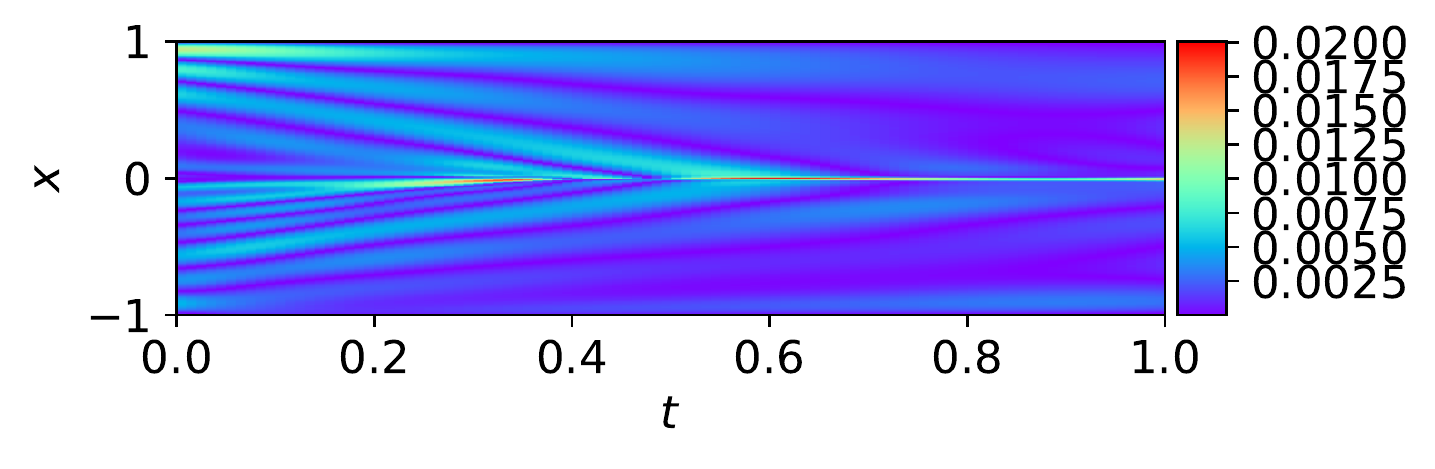}}
				\end{minipage}
				&\begin{minipage}[b]{0.6\columnwidth}
					\centering
					\raisebox{-.5\height}{
						\includegraphics[width=\linewidth]{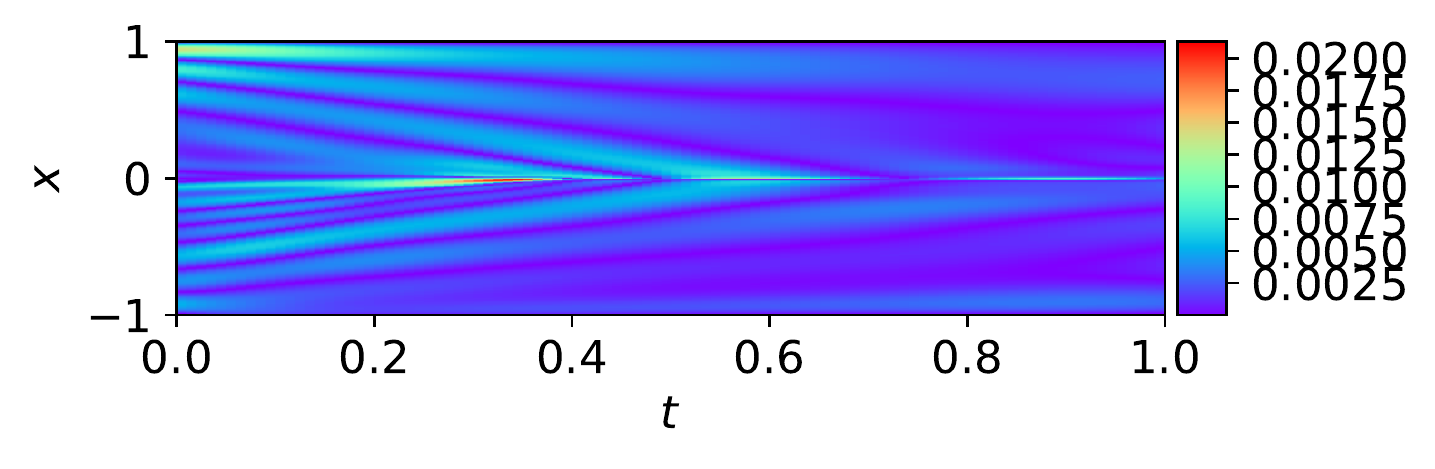}}
				\end{minipage}
				&\begin{minipage}[b]{0.6\columnwidth}
					\centering
					\raisebox{-.5\height}{
						\includegraphics[width=\linewidth]{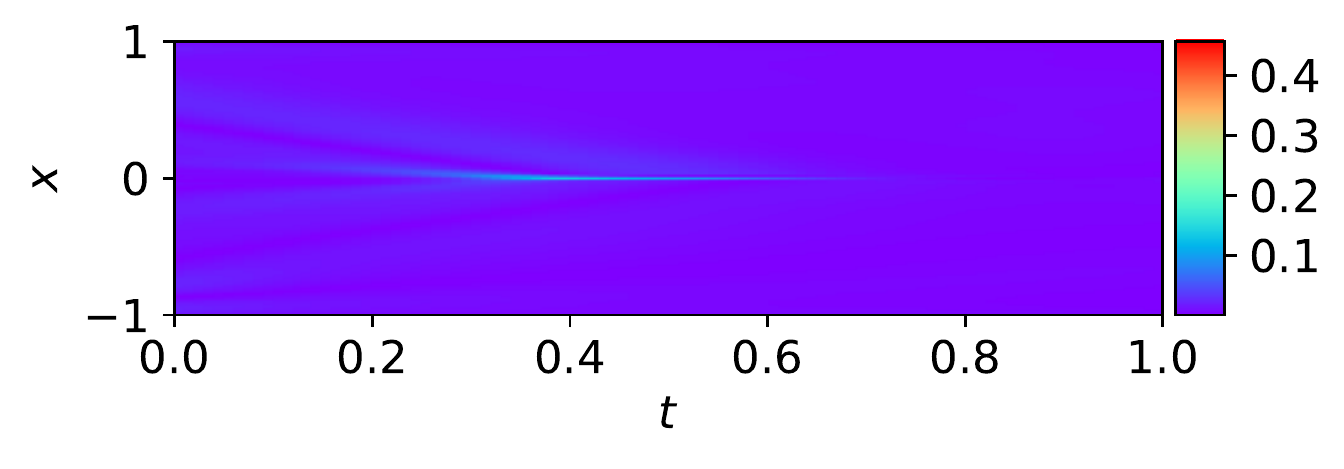}}
				\end{minipage}
				\\ 
				\hline
				\makecell[c]{\rotatebox{90}{{\small Schrodinger}}}&\begin{minipage}[b]{0.6\columnwidth}
					\centering
					\raisebox{-.5\height}{
						\includegraphics[width=\linewidth]{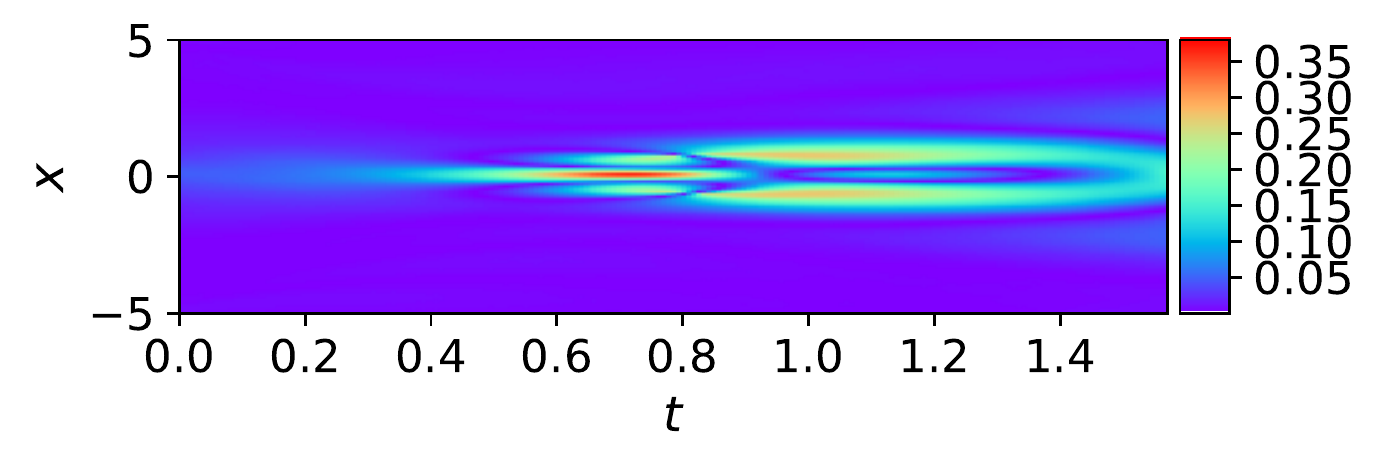}}
				\end{minipage}
				&\begin{minipage}[b]{0.6\columnwidth}
					\centering
					\raisebox{-.5\height}{
						\includegraphics[width=\linewidth]{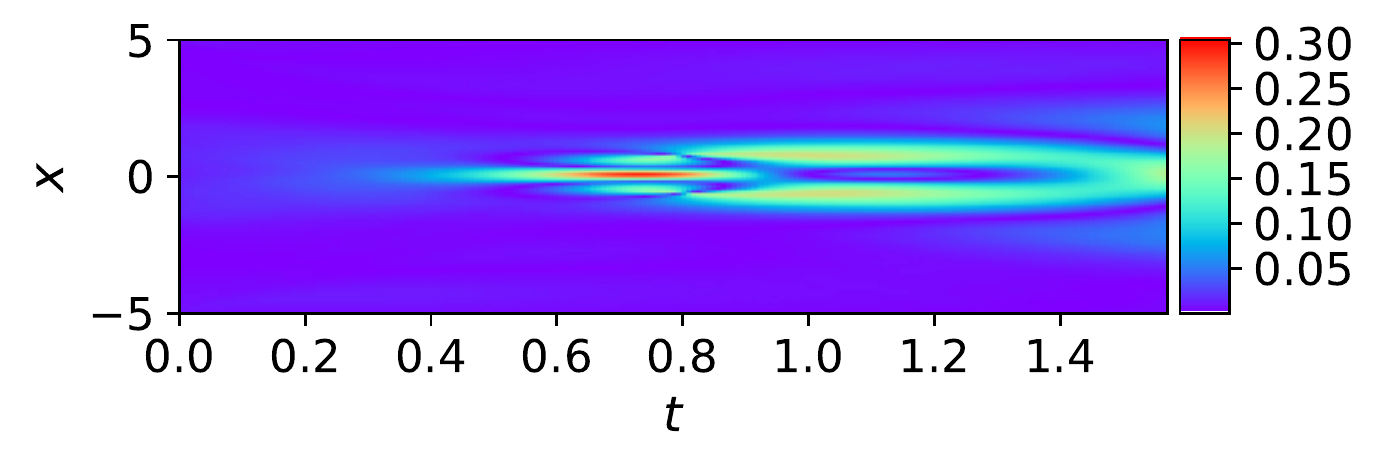}}
				\end{minipage}
				&\begin{minipage}[b]{0.6\columnwidth}
					\centering
					\raisebox{-.5\height}{
						\includegraphics[width=\linewidth]{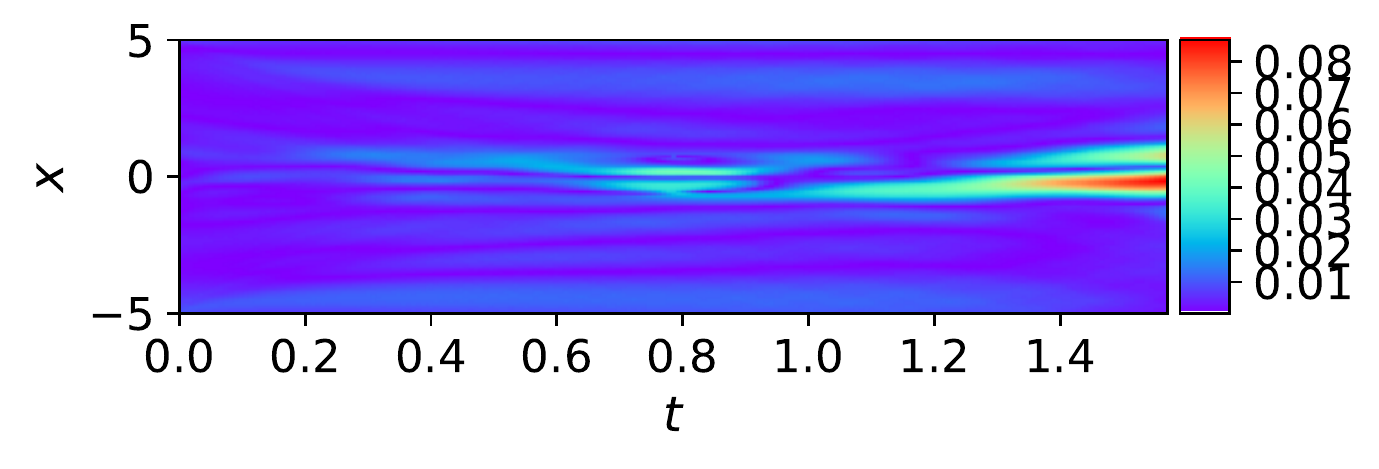}}
				\end{minipage}
				&\begin{minipage}[b]{0.6\columnwidth}
					\centering
					\raisebox{-.5\height}{
						\includegraphics[width=\linewidth]{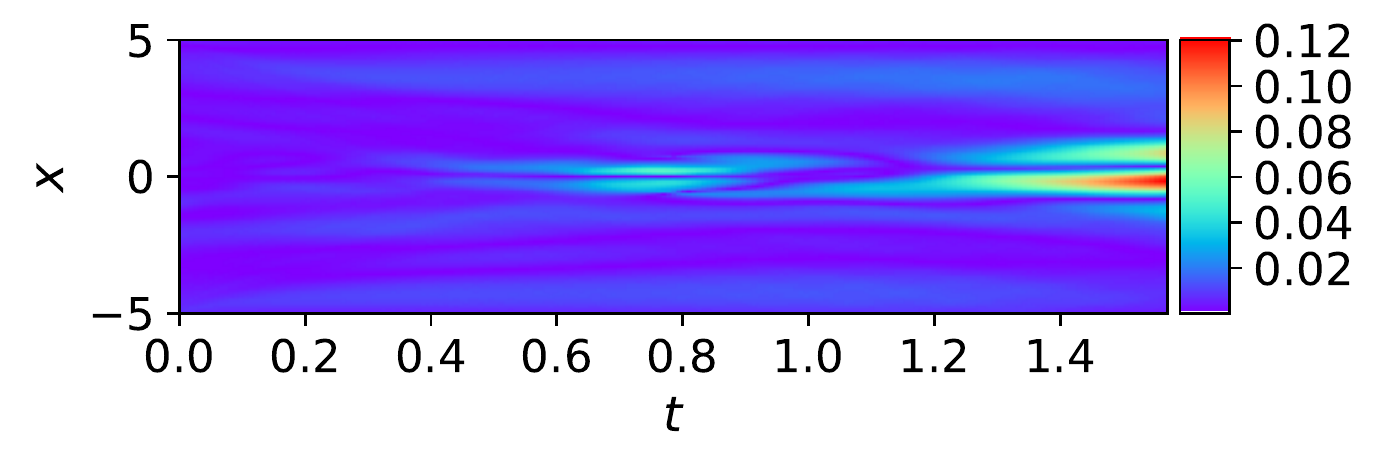}}
				\end{minipage}
				&\begin{minipage}[b]{0.6\columnwidth}
					\centering
					\raisebox{-.5\height}{
						\includegraphics[width=\linewidth]{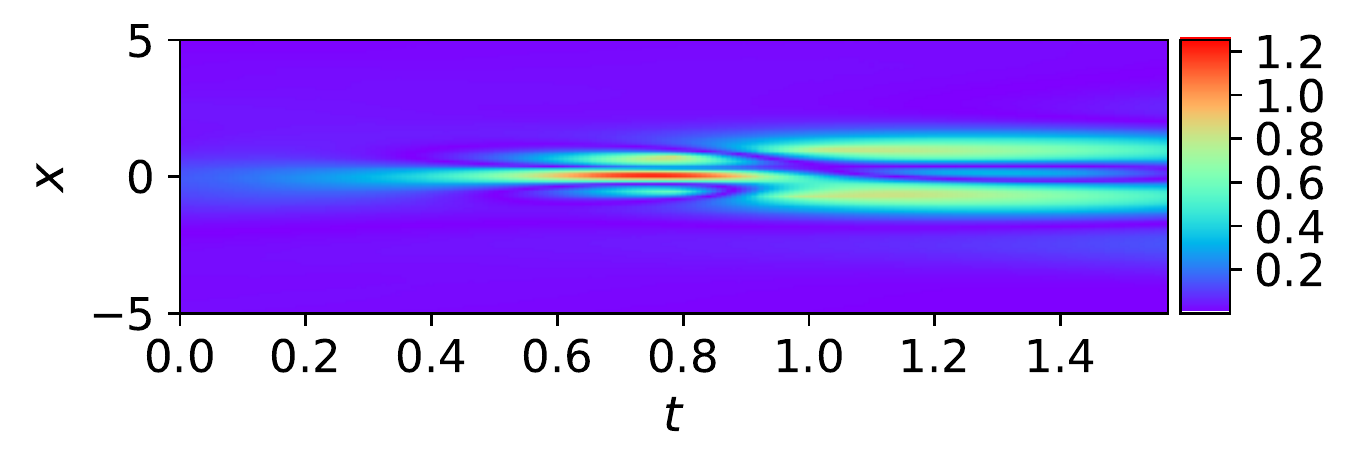}}
				\end{minipage}
				\\ 
				\hline
				\makecell[c]{\rotatebox{90}{{\small Helmholtz}}}&\begin{minipage}[b]{0.6\columnwidth}
					\centering
					\raisebox{-.5\height}{
						\includegraphics[width=\linewidth]{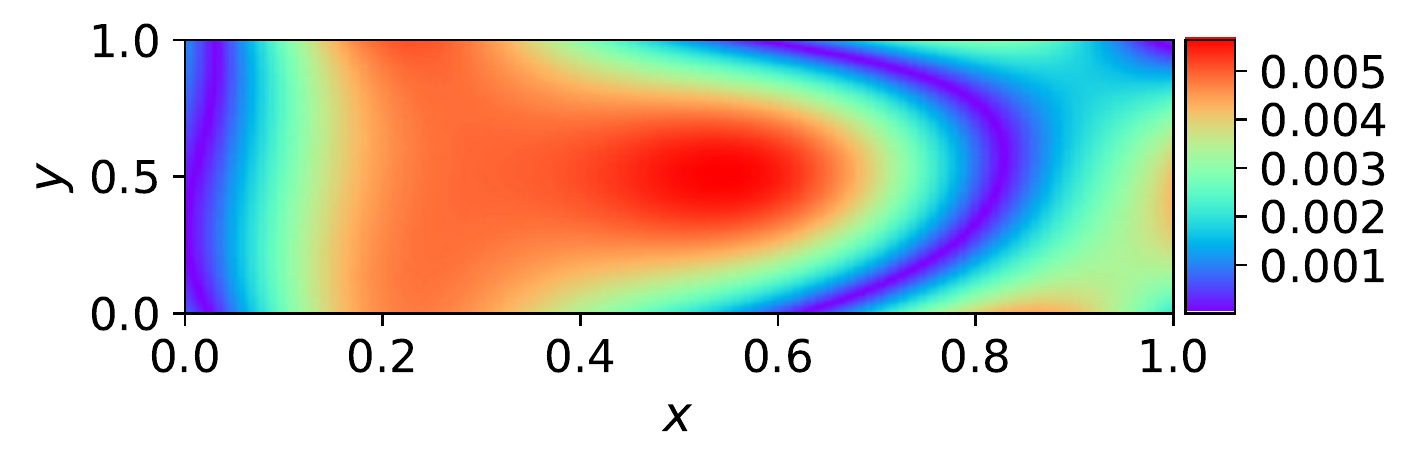}}
				\end{minipage}
				&\begin{minipage}[b]{0.6\columnwidth}
					\centering
					\raisebox{-.5\height}{
						\includegraphics[width=\linewidth]{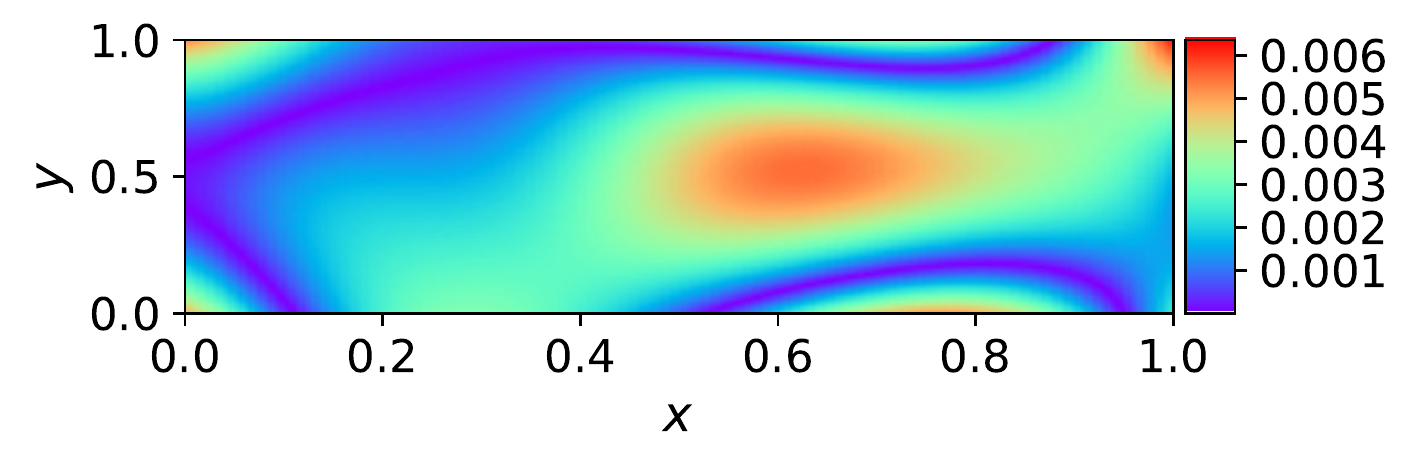}}
				\end{minipage}
				& \begin{minipage}[b]{0.6\columnwidth}
					\centering
					\raisebox{-.5\height}{
						\includegraphics[width=\linewidth]{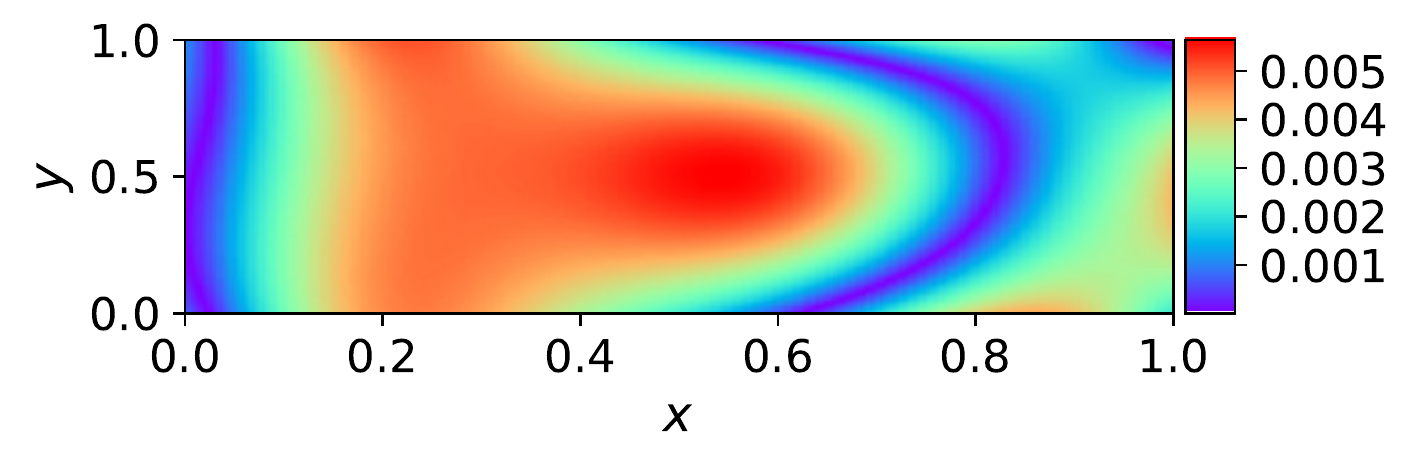}}
				\end{minipage}
				&\begin{minipage}[b]{0.6\columnwidth}
					\centering
					\raisebox{-.5\height}{
						\includegraphics[width=\linewidth]{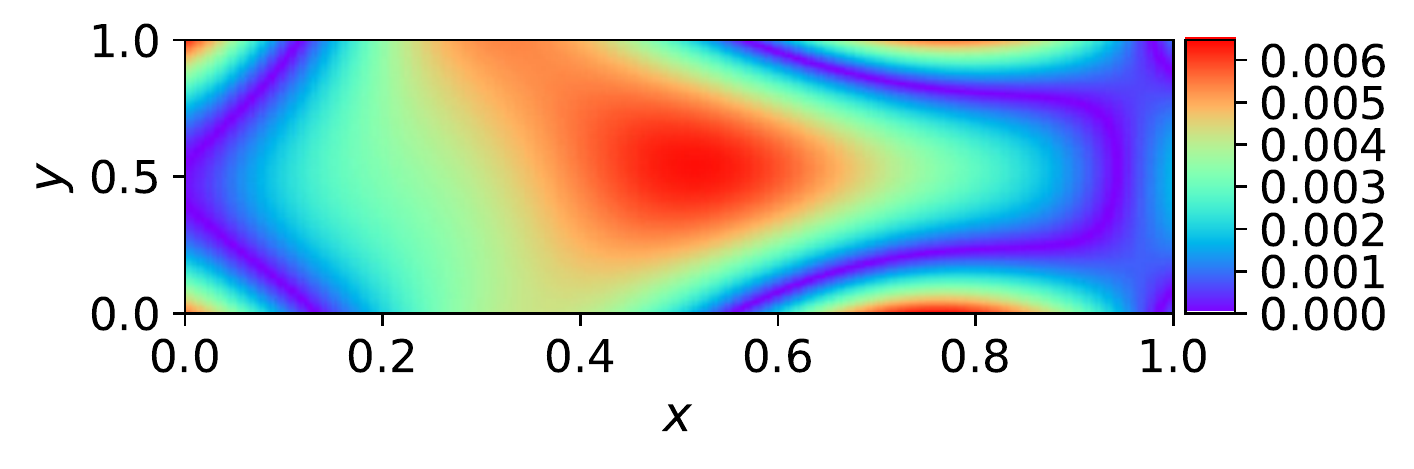}}
				\end{minipage}
				&\begin{minipage}[b]{0.6\columnwidth}
					\centering
					\raisebox{-.5\height}{
						\includegraphics[width=\linewidth]{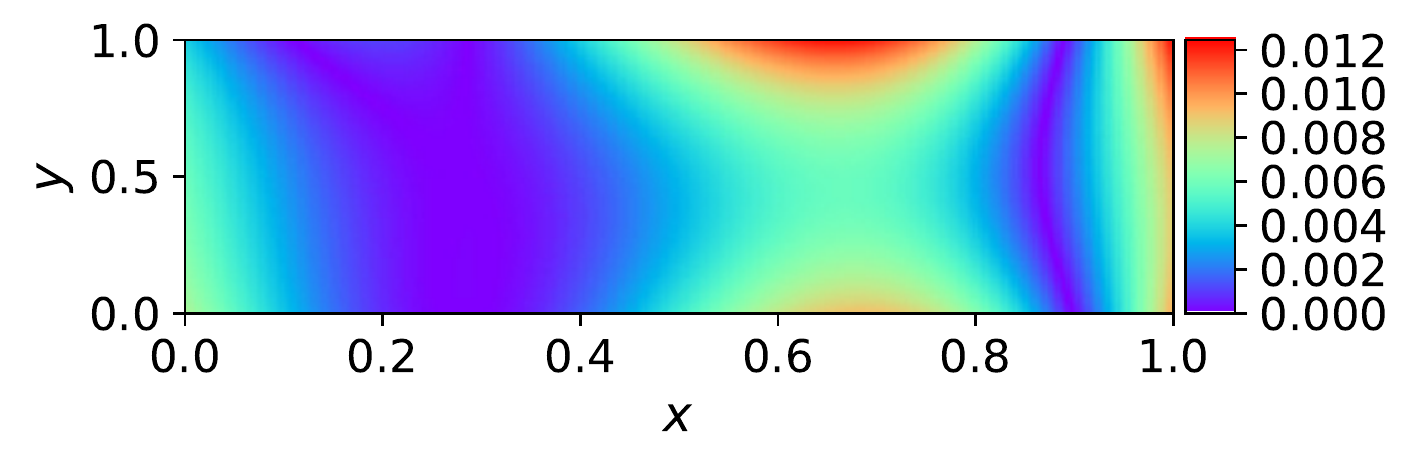}}
				\end{minipage}
				\\ 
				\hline
				\makecell[c]{\rotatebox{90}{{\small Poisson}}}&\begin{minipage}[b]{0.6\columnwidth}
					\centering
					\raisebox{-.5\height}{
						\includegraphics[width=\linewidth]{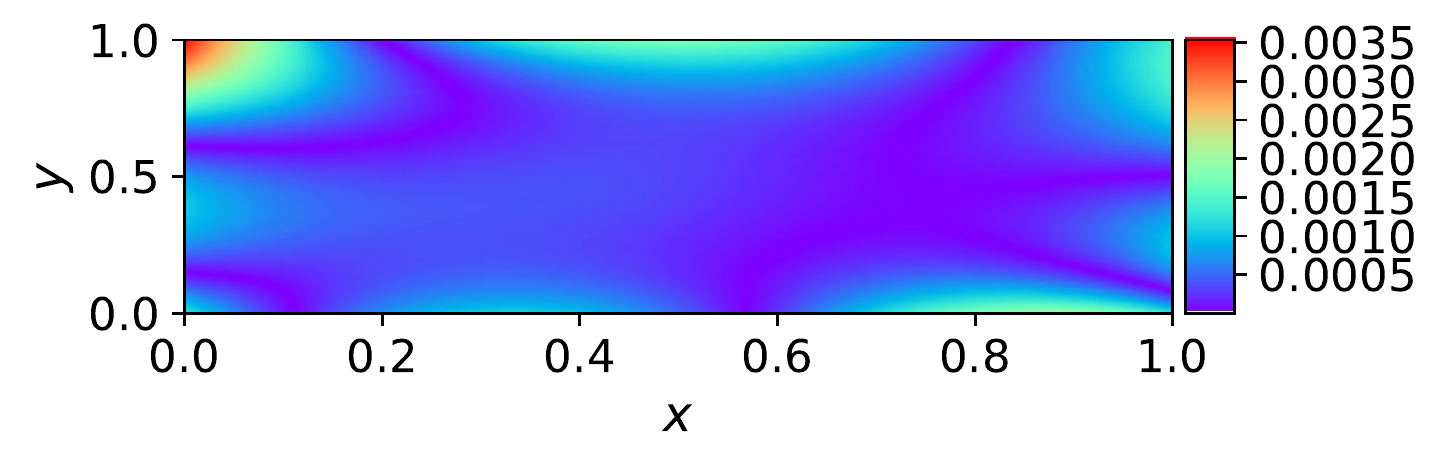}}
				\end{minipage}
				&\begin{minipage}[b]{0.6\columnwidth}
					\centering
					\raisebox{-.5\height}{
						\includegraphics[width=\linewidth]{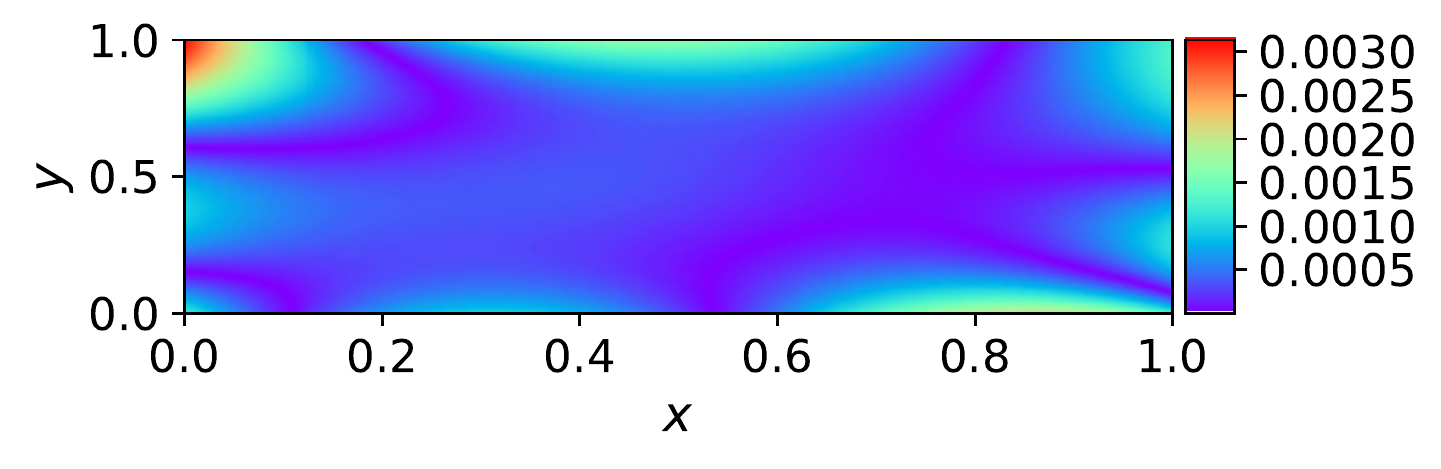}}
				\end{minipage}
				&\begin{minipage}[b]{0.6\columnwidth}
					\centering
					\raisebox{-.5\height}{
						\includegraphics[width=\linewidth]{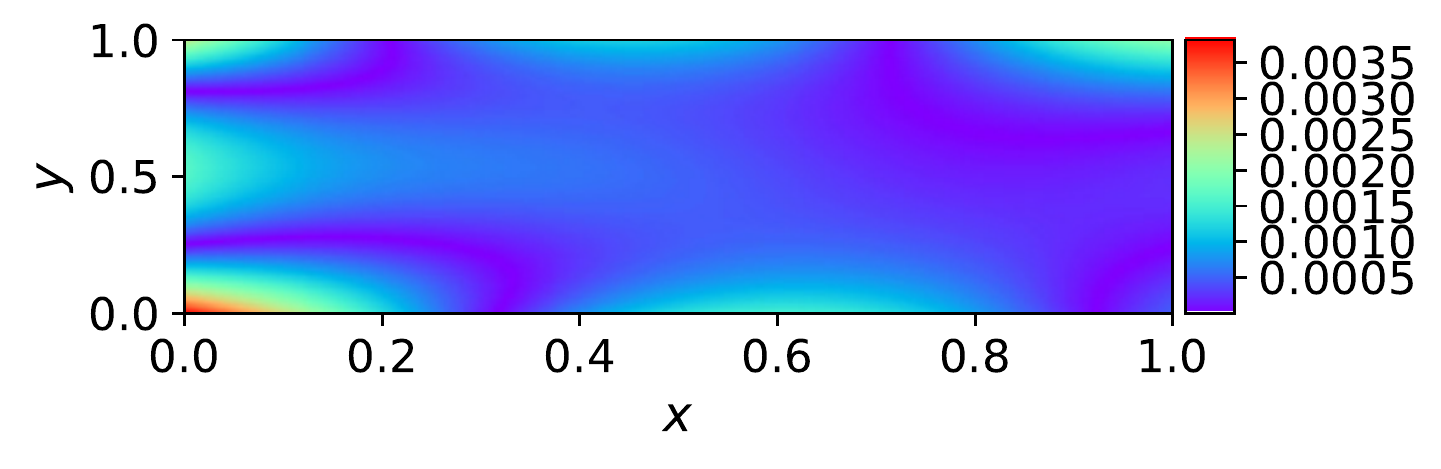}}
				\end{minipage}
				&\begin{minipage}[b]{0.6\columnwidth}
					\centering
					\raisebox{-.5\height}{
						\includegraphics[width=\linewidth]{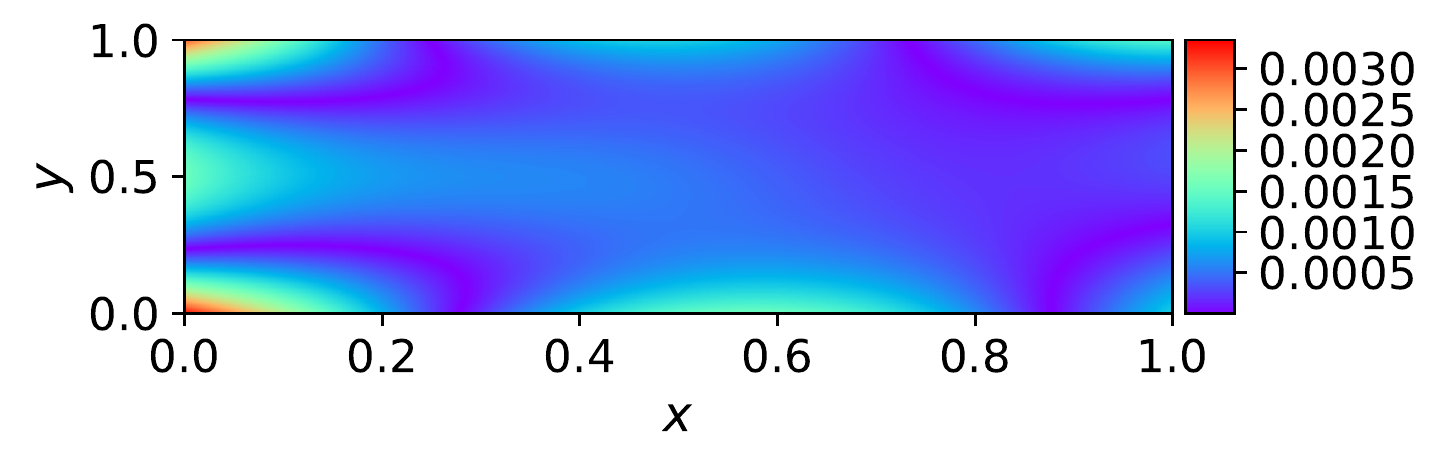}}
				\end{minipage}
				&\begin{minipage}[b]{0.6\columnwidth}
					\centering
					\raisebox{-.5\height}{
						\includegraphics[width=\linewidth]{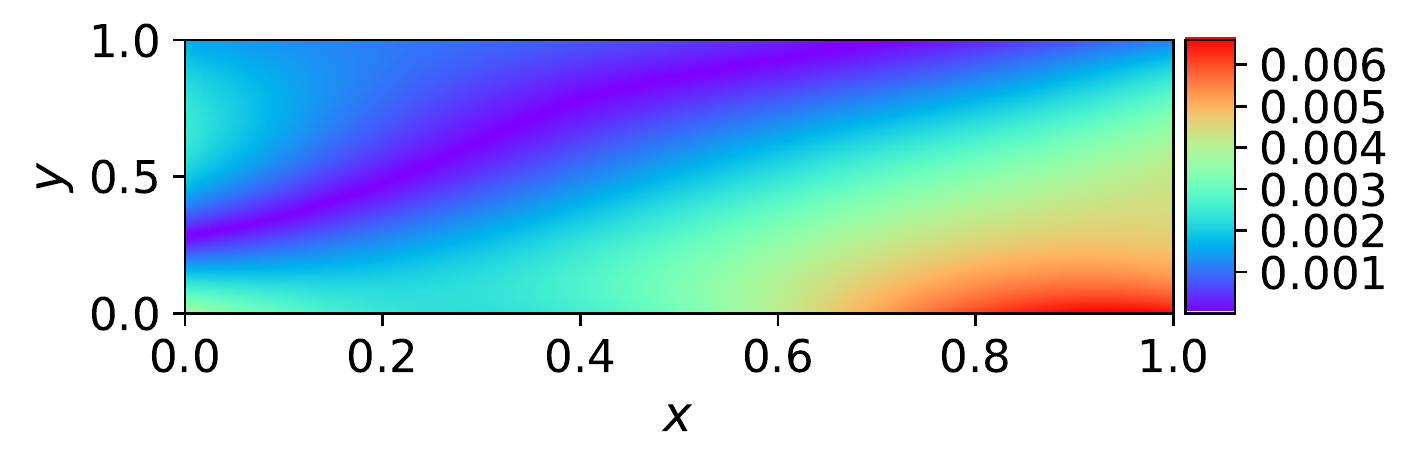}}
				\end{minipage}
				\\ 
				\hline
				\makecell[c]{\rotatebox{90}{{\small Heat}}}&\begin{minipage}[b]{0.6\columnwidth}
					\centering
					\raisebox{-.5\height}{
						\includegraphics[width=\linewidth]{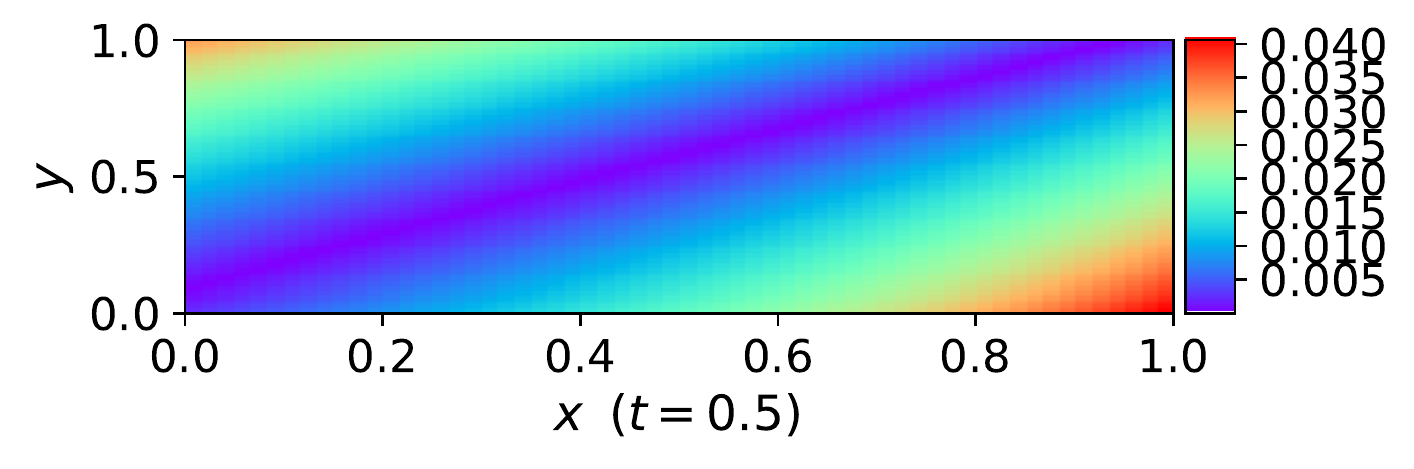}}
				\end{minipage}
				&\begin{minipage}[b]{0.6\columnwidth}
					\centering
					\raisebox{-.5\height}{
						\includegraphics[width=\linewidth]{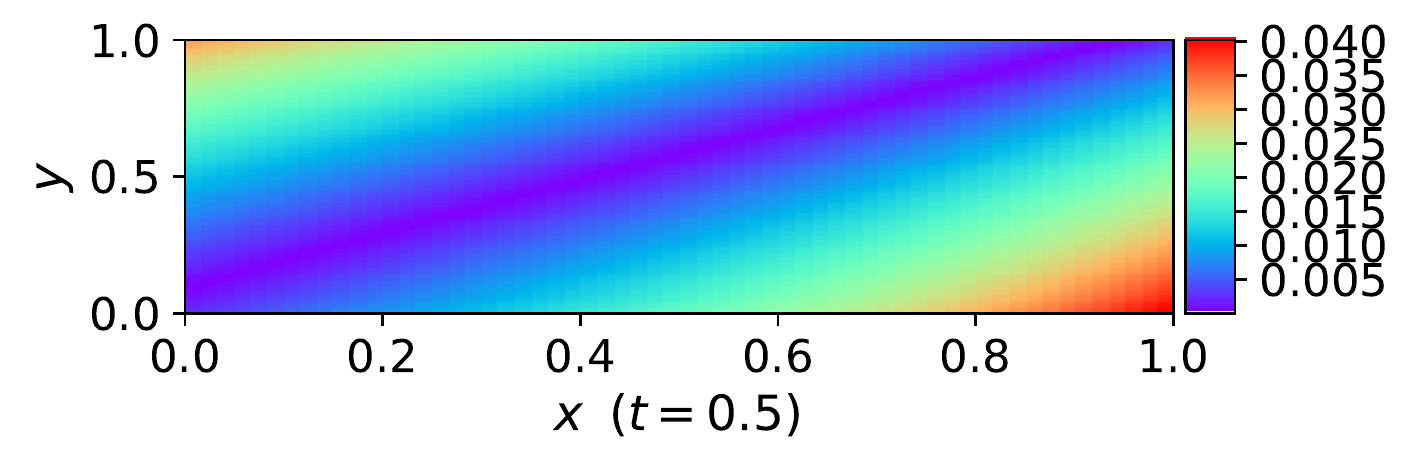}}
				\end{minipage}
				&\begin{minipage}[b]{0.6\columnwidth}
					\centering
					\raisebox{-.5\height}{
						\includegraphics[width=\linewidth]{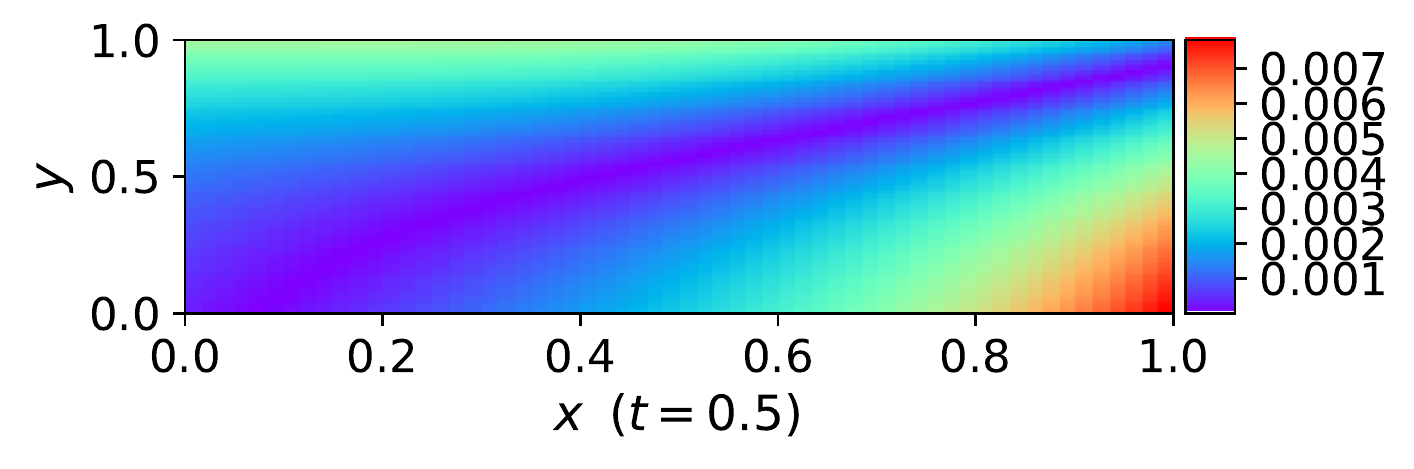}}
				\end{minipage}
				&\begin{minipage}[b]{0.6\columnwidth}
					\centering
					\raisebox{-.5\height}{
						\includegraphics[width=\linewidth]{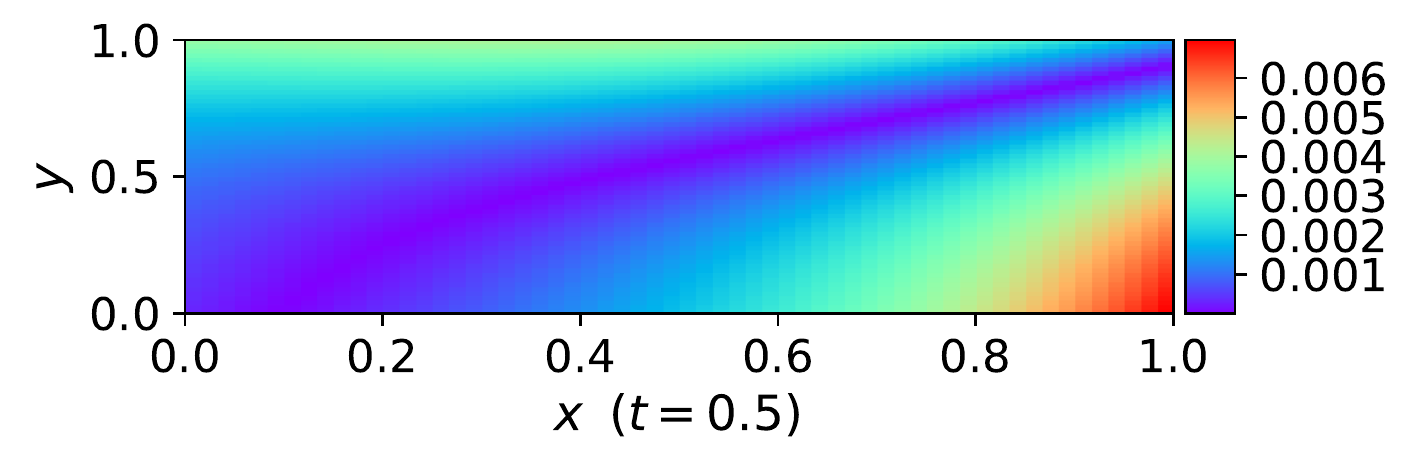}}
				\end{minipage}
				&\begin{minipage}[b]{0.6\columnwidth}
					\centering
					\raisebox{-.5\height}{
						\includegraphics[width=\linewidth]{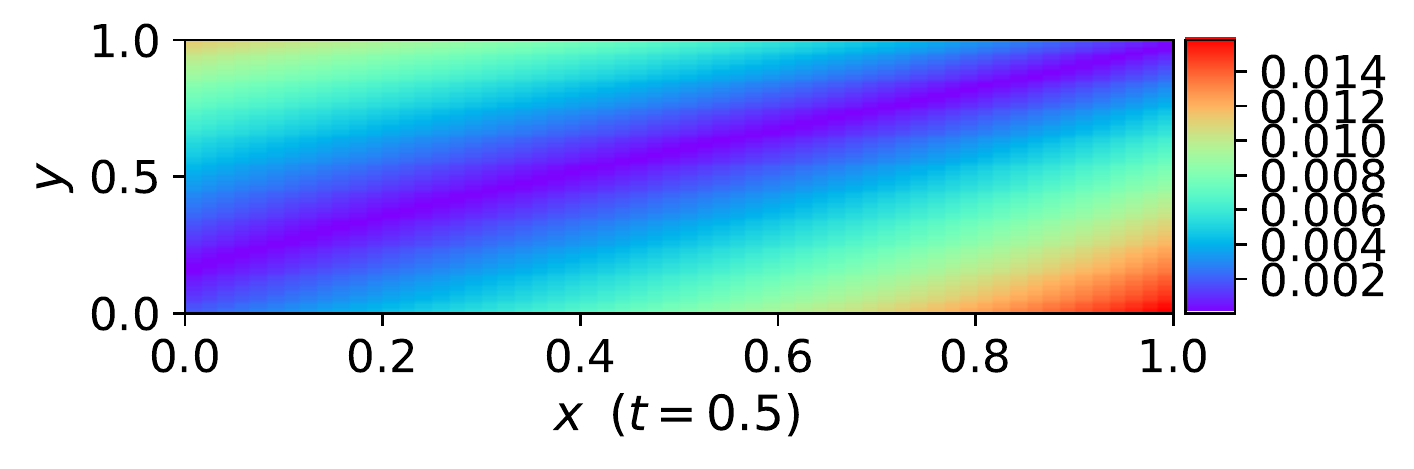}}
				\end{minipage}
				\\ 
				\hline
			\end{tabular}
		\end{threeparttable}\label{tab:heat}
	}
\end{table*}

All experiments are processed in the DELL$^{\textregistered}$ PowerEdge$^{\textregistered}$ T640 Tower Server with two Intel$^{\textregistered}$ Xeon$^{\textregistered}$ 20-core processors, 128 GB RAM and a NVIDIA$^{\textregistered}$ Tesla$^{\textregistered}$ V100 16GB GPU. We select six well-known PDEs as the testing problems, including Burgers equation, Poisson equation, Helmholtz equation, Schrodinger equation, High-dimensional (HD) Poisson equation and Heat equation ({\it cf.} Tab. \ref{tab:pde}).
We consider the following models for numerically solving these PDEs, including PINN, PINN+PW, GA-PINN and GA-PINN+PW. Given a testing data $\{({\bf x}^{(l)},u^{(l)})\}_{l=1}^L$, we adopt the normalized root mean square error (NRMSE) as the criterion of testing performance:   
\begin{equation*}
	{\rm NRMSE}:= \sqrt{\frac{\sum_{l=1}^L \big\| \widehat{u}({\bf x}^{(l)})  - u^{(l)}\big\|^2 }{\sum_{l=1}^L \big\|u^{(l)}\big\|^2}}. 
\end{equation*}

For the sake of fairness, in each testing PDE problem, PINN, PINN+PW and the generators of GA-PINN and GA-PINN+PW share the same structure and initial weights that are obtained by using the Xavier initialization \citep{he2015delving}. Their objective functions are minimized by using Adam optimization algorithm \citep{kingma2014adam}. Following the setting of PINNs given by \citet{raissi2019physics}, the Adam method is of the batch form, {\it i.e.,} all samples are used to compute the objective functions at each iteration. Different from the ordinary supervised learning tasks, where the labeled samples have contained the relationship between the input and the output, the training of PINNs is achieved by minimizing the PI loss ${\rm L}_{\rm PINN}$ that is computed on the points taken from the domain and its boundaries. Therefore, when using the Adam method with a small mini-batch, the incomplete physics information provided by the PI loss ${\rm L}_{\rm PINN}$ computed on the insufficient points could misguide the updating direction of network weights. Moreover, since the size of labeled samples is small, the mini-batch Adam method is not be used to optimize the generative loss ${\rm L}_G$ and the discriminative loss ${\rm L}_D$ as well. 

The termination condition of training these models is that the PI loss ${\rm L}_{\rm PINN}$ computed on the sample points reaches the pre-defined level, and meanwhile we use ${\rm L}_{\rm PINN}$ as 
the criterion of training performance as well. The reason why to set such a termination condition is because a well-trained network should be able to recover the complete physics information, and thus ${\rm L}_{\rm PINN}$ actually acts as a criterion to evaluate the completeness of physics information. The choices of the hyperparameters are listed in Tab. \ref{tab:hyperparameter}.

As a comparison, we also adopt the deep Galerkin method (DGM) \citep{sirignano2018dgm} to solve these testing PDEs. As addressed in Section \ref{sec:related}, since DGM actually is a PINN with the SGD method, DGM share the same learning rate $\eta_P$ and network structure with the PINN for each testing PDE problem. In DGM, the sizes of the mini-batches for ${\rm L}_f$ and ${\rm L}_b$ are both set to be $256$ so as to maintain the relatively complete physics information and the points are randomly drawn from the domain and its boundaries at each iteration, respectively.
The termination condition of training is set to reach the pre-defined iteration epoch ({\it cf.} Tab. \ref{tab:NRMSE}).

\begin{figure}[htbp]
	\centering 
	\begin{minipage}[t]{0.32\linewidth}
		\centerline{\includegraphics[width=0.9\linewidth]{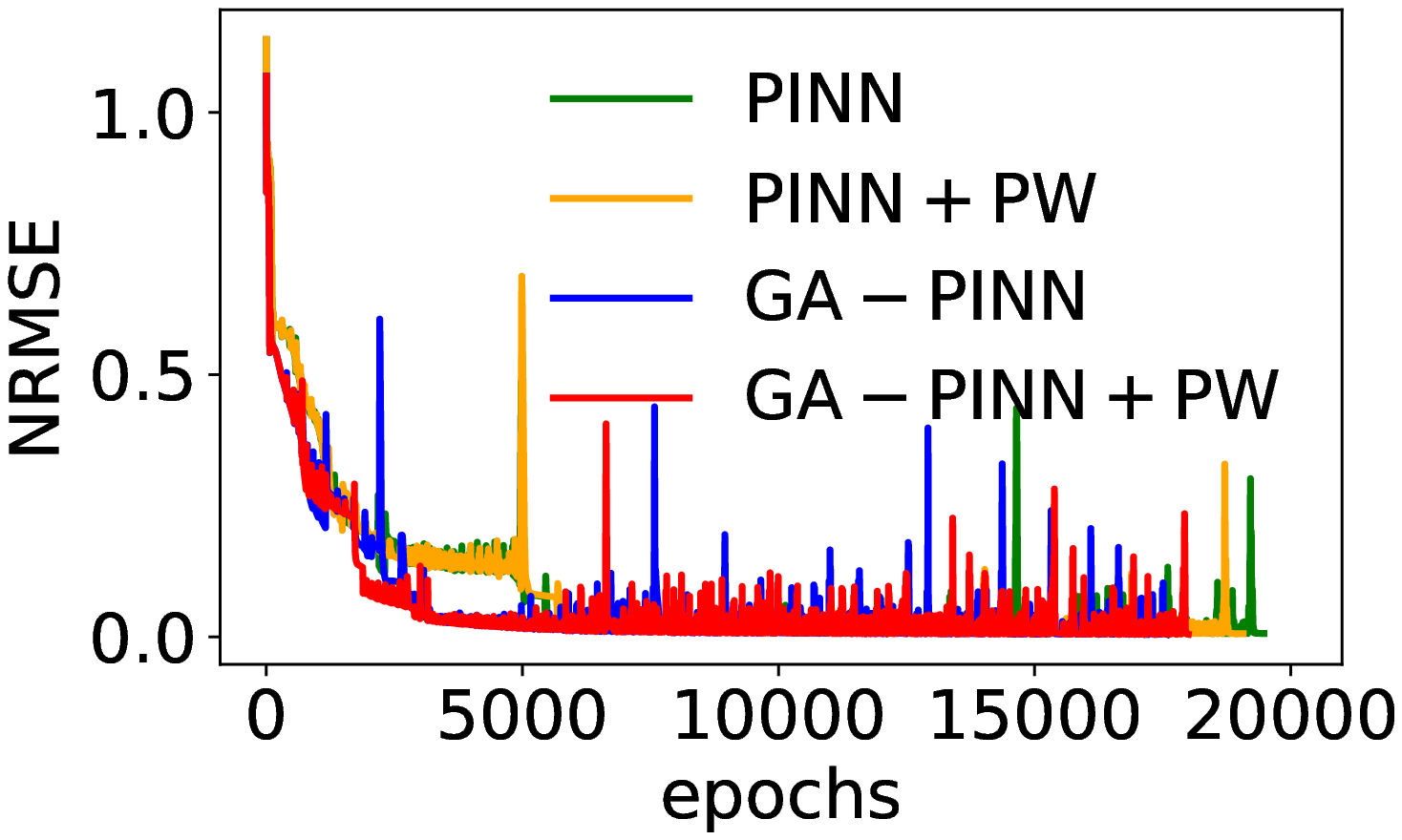}}
		\centerline{(a) Burgers}
	\end{minipage}
	\begin{minipage}[t]{0.32\linewidth}
		\centerline{\includegraphics[width=0.9\linewidth]{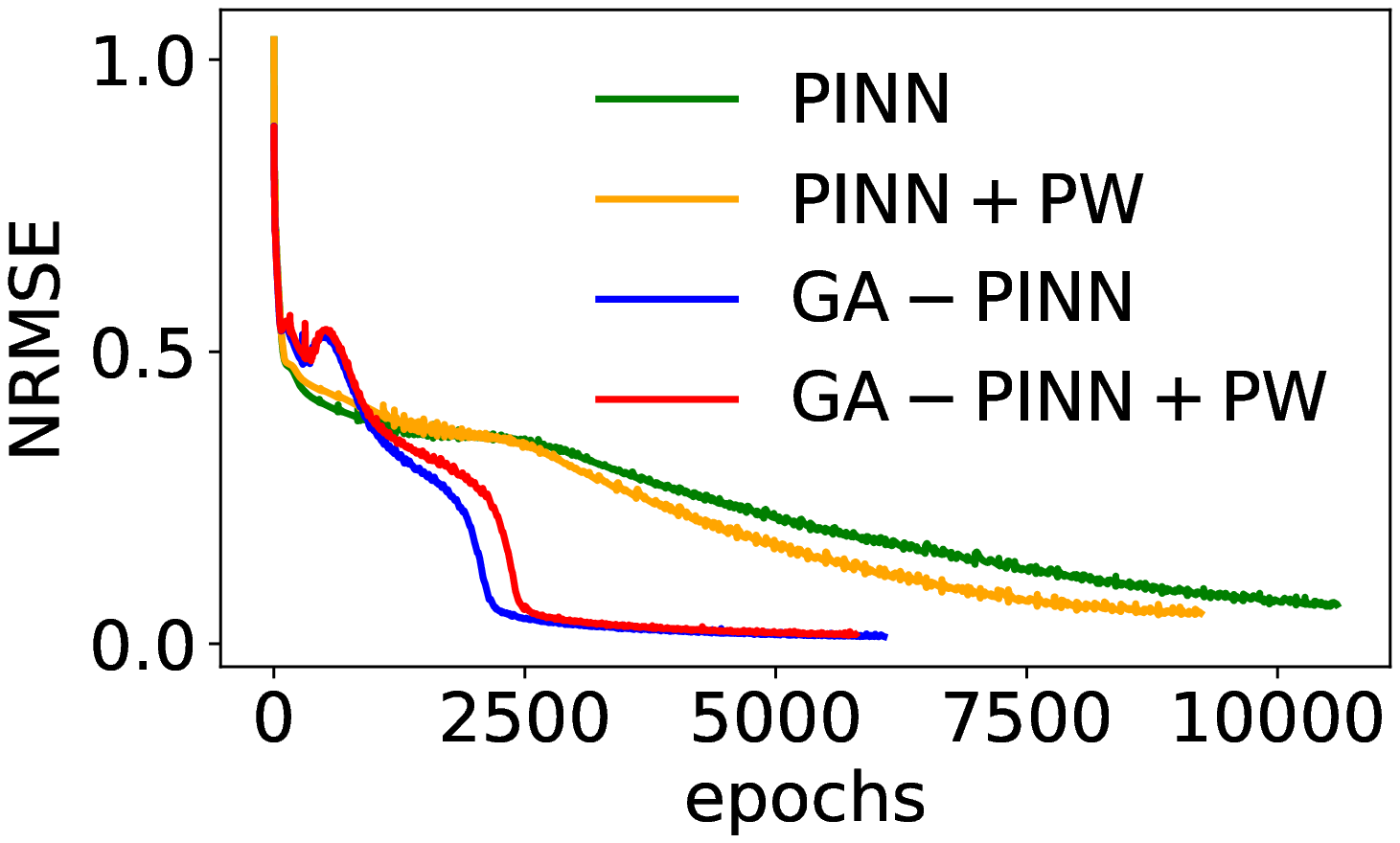}}
		\centerline{(b) Schrodinger}
	\end{minipage}
	\begin{minipage}[t]{0.32\linewidth}
		\centerline{\includegraphics[width=0.9\linewidth]{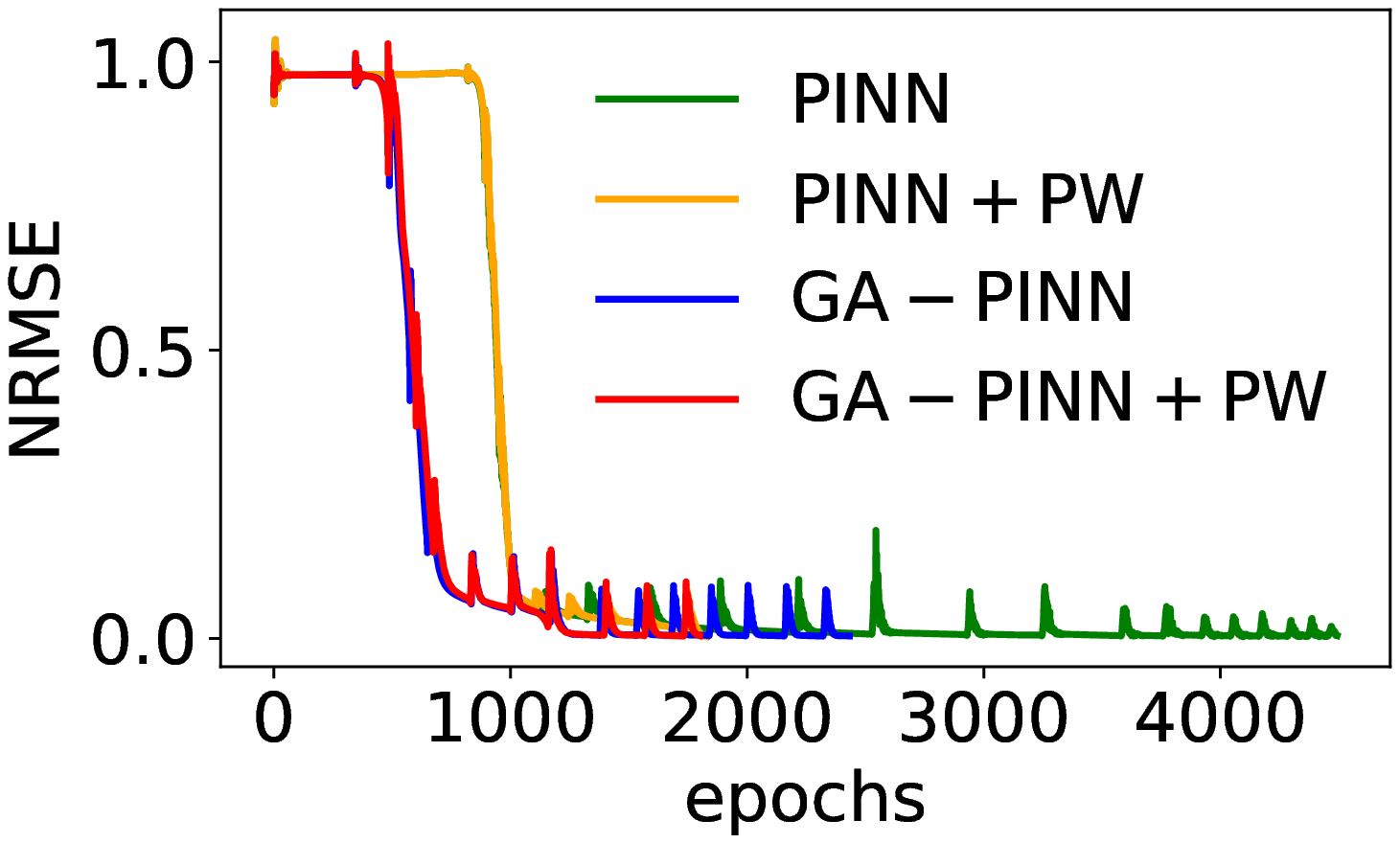}}
		\centerline{(c) Helmholtz}
	\end{minipage}
	\begin{minipage}[t]{0.32\linewidth}
		\centerline{\includegraphics[width=0.9\linewidth]{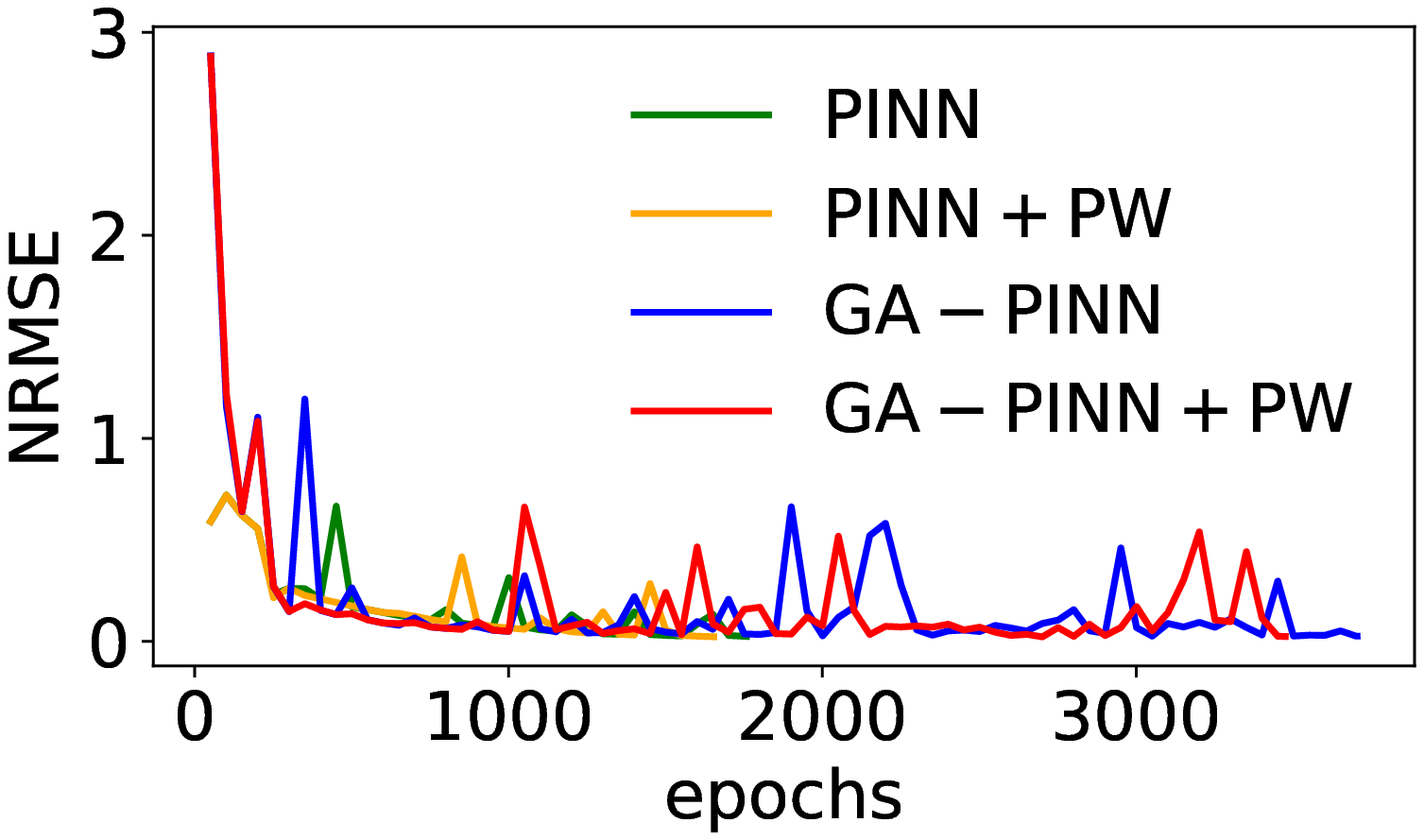}}
		\centerline{(d) Poisson}
	\end{minipage}
	\begin{minipage}[t]{0.32\linewidth}
		\centerline{\includegraphics[width=0.9\linewidth]{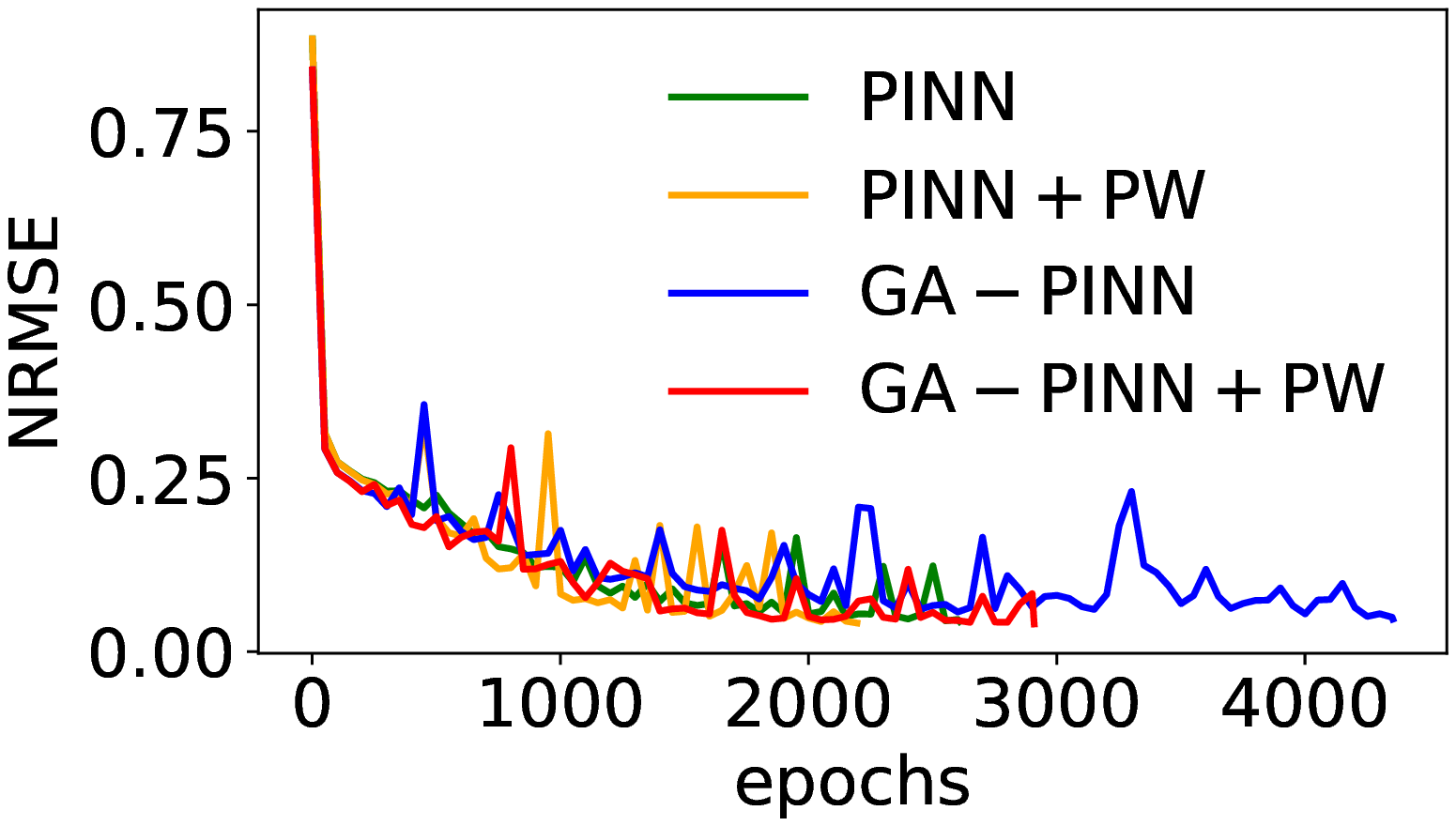}}
		\centerline{(e) HD Poisson}
	\end{minipage}
	\begin{minipage}[t]{0.32\linewidth}
		\centerline{\includegraphics[width=0.9\linewidth]{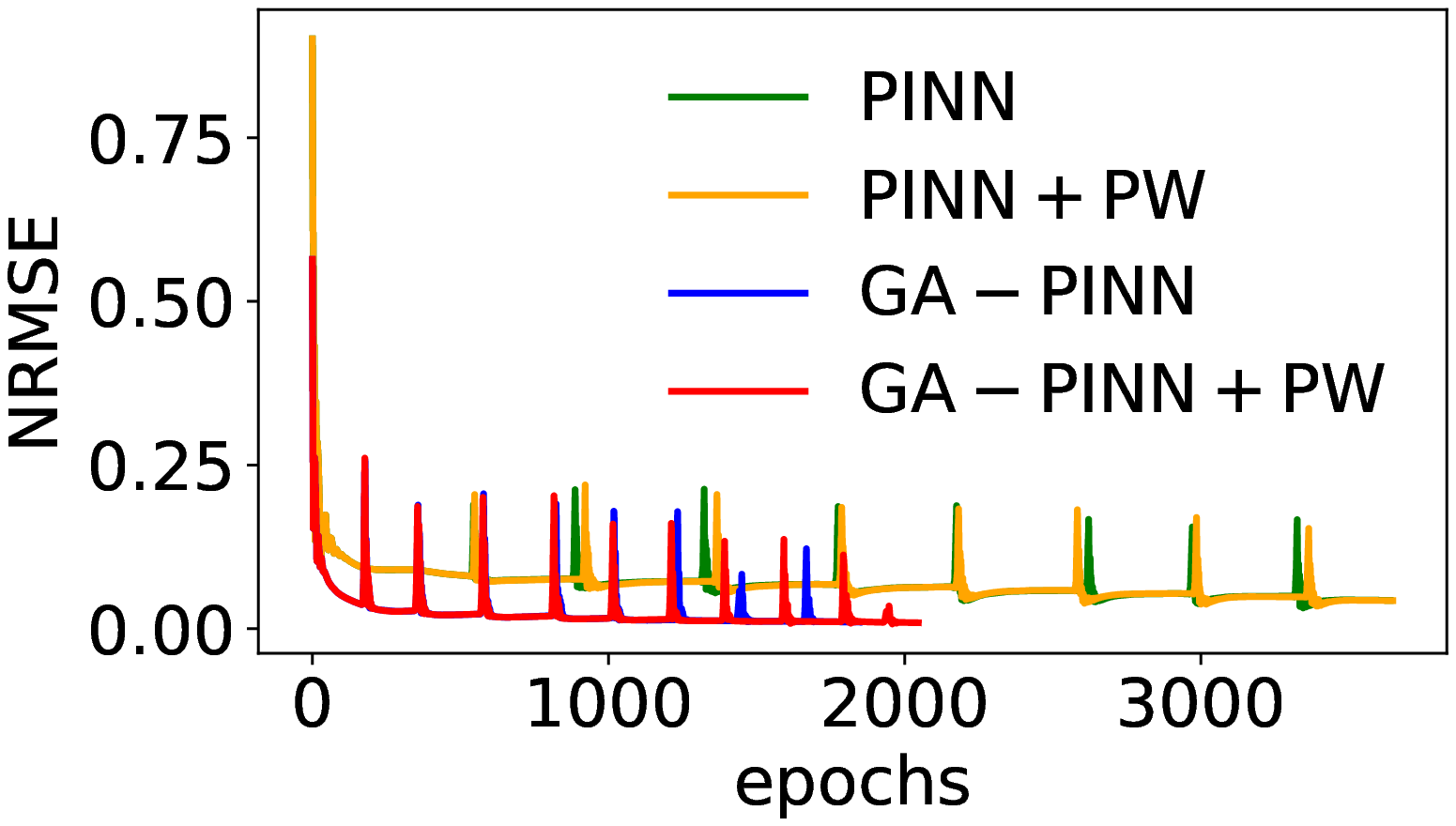}}
		\centerline{(f) Heat}
	\end{minipage}
	\caption{The curves of testing error ({\rm NRMSE}) for different testing PDEs.
	} 
	\label{fig:NRMSE}
\end{figure}

\subsection{Experimental Results for PW Method}

Here, we consider the effect of the PW method in the training process of PINNs. For each testing PDE problem, we first study the minimization process of ${\rm L}_f$ and ${\rm L}_b$ with or without the PW method, and then analyze the training process of PINN and PINN+PW. As shown in the first two columns of Tab. \ref{tab:pw}, the PW method can speed up the minimization process of ${\rm L}_f$ and ${\rm L}_b$ when they have sharp fluctuation, but do not evidently influence the smooth minimization process of the objective functions, {\it e.g.,} the ${\rm L}_b$ of Heat and the ${\rm L}_f$s of Burgers, Helmholtz, Schrodinger, HD Poisson and Heat. Therefore, in the following experiments, we only apply the PW method to speed up the minimization process of ${\rm L}_b$ in the five equations, and the PW method is used to minimize both of ${\rm L}_f$ and ${\rm L}_b$ in the process of training PINN+PW for Poisson equation.

As shown in the last two columns of Tab. \ref{tab:pw} and the first two rows of Tab. \ref{tab:NRMSE}, PINN+PW not only reach the training termination condition earlier than PINN, but also provides a lower testing performance (NRMSE) in most cases. It implies that the introduction of the PW method can evidently improve the performance of PINNs. Interestingly, we also find that the effect of the PW method is not evident to the minimization of ${\rm L}_f$ in Poisson equation, but the training process of PINN+PW is much faster than that of PINN with a lower testing NRMSE.


\subsection{Experimental Results for GA-PINNs}

GA-PINNs aim to use a small size of (usually very few) labeled samples, which are the exact solutions (or their high-accuracy approximations) to a PDE at some discrete points in the domain $\Omega$, to refine the numerical solutions to the PDE provided by PINNs. At each iteration of training GA-PINN, after minimizing the discriminative loss ${\rm L}_D$ and the generative loss ${\rm L}_G$, the PI loss ${\rm L}_{\rm PINN}$ (or the weighted PI loss ${\rm L}^{\rm PW}_{\rm PINN}$) is minimized to fine-tune the generator weights that could be misguided by the discriminator trained based on insufficient labeled samples. To further examine the effectiveness of the PW method, we also compare the performance of GA-PINN with that of GA-PINN+PW ({\it cf.} Alg. \ref{alg:ga-pinn}).

Figure \ref{fig:NRMSE} and Tables \ref{tab:performance}$\sim$\ref{tab:heat} show that the introduction of the GA mechanism can effectively exploit the small size of labeled samples to speed up the training process and improve the accuracy of the numerical solutions. However, the PW method sometimes will bring a negative effect to the process of training GA-PINNs in some complicated PDEs such as Burgers equation and Schrodinger equation. The reason is that the introduction of the PW method will influence the minimax game training, which could become fragile in the case that the PDEs are complicated but the labeled samples are not sufficient. In addition, PINN+PW performs comparably with GA-PINN (or GA-PINN+PW) in the relatively simple problems, such as Helmholtz, Poisson and HD Poisson. This finding implies that PINN+PW is a good candidate for numerically solving simple PDEs. Moreover, benefiting from the randomness of the SGD method, DGM has a good performance in HD Poisson but needs a relatively large epoch number and mini-batch size.


\section{Conclusion}

In this paper, we consider the numerical solutions to PDEs in the situation that the exact solutions (or their high-accuracy approximations) to PDEs are available at a small amount of (usually very few) discrete points in the domain. This situation is common in practice, but the mechanism of PINNs is unsuitable (at least cannot be directly applied) to this situation ({\it cf.} Section \ref{sec:pinn}). To overcome this limitation, we integrate the GA mechanism with the structure of PINNs to form the proposed GA-PINNs. 

The GA-PINN is composed of two sub-networks: a generator and a discriminator. The generator, with the same structure as PINNs, produces the numerical solution $\widehat{u}({\bf x})$ to a PDE associated with the input point ${\bf x}\in\Omega$. The discriminator aims to identify whether the pair $({\bf x}, \widehat{u}({\bf x}))$ is a real exact solution to the PDE. The training process of GA-PINNs is achieved by using a small size of labeled samples $\{({\bf x}_T^{(j)}, u_T^{(j)})\}_{j=1}^J$. Each iteration of training GA-PINNs contains two stages: one is the minimax game between the generator and the discriminator and the other is the minimization of ${\rm L}_{\rm PINN}$ (or ${\rm L}^{\rm PW}_{\rm PINN}$) to refine the generator  after the minimax game training. The goal of the second stage is to fine-tune the generator weights influenced by the insufficient labeled samples. Taking advantage of the GA mechanism, GA-PINNs are able to effectively exploit the insufficient labeled samples to improve the accuracy of numerical solutions to PDEs. The experimental results support the effectiveness of GA-PINNs and show that GA-PINNs outperform PINNs in these testing PDE problems.

Since the efficiency of training PINNs often becomes low especially for some complicated PDEs, we also propose the PW method to speed up the process of training PINNs. This method splits the training process into two stages: one is to promote the network to reach a relatively stable status by increasing the EL-point weights; and the other is to improve the network performance by increasing the HL-point weights. The numerical experiments show that the PW method can evidently improve the efficiency of training PINNs with a lower testing performance. Interestingly, the PW method cannot influence the smooth minimization process, but only speeds up the minimization process that has the sharp fluctuation. Therefore, the PW method has high applicability in practice. Moreover, we also introduce the PW method into the process of training GA-PINNs (called GA-PINN+PW accordingly) ({\it cf.} Alg. \ref{alg:ga-pinn}), and the experimental results demonstrate that GA-PINN+PW performs better than GA-PINN in most cases. In the further works, we will consider the feasibility of the PW method in training other deep learning models such as AlexNet, VGGNet and ResNet, and use GA-PINNs to numerically solve some important PDEs such as Navier-Stokes equations.


%



\bibliography{ga-pinn}
\bibliographystyle{icml2022}

\end{document}